\documentclass[acmtog]{acmart}

\graphicspath{{./figures/}}
\usepackage{colortbl}
\usepackage[acronym,nomain]{glossaries}
\glsdisablehyper
\usepackage{makecell}
\setcellgapes{5pt}

\providecommand{\abs}[1]{\lvert#1\rvert}
\DeclareMathOperator*{\argmin}{argmin}
\DeclareMathOperator{\E}{\mathbb{E}}
\DeclareMathOperator{\Var}{Var}
\DeclareMathOperator{\Cov}{Cov}
\newcommand{\best}[1]{\cellcolor{red!40}\textbf{#1}}
\newcommand{\second}[1]{\cellcolor{orange!40}\textit{#1}}

\newacronym{hdr}{HDR}{high dynamic range}
\newacronym{spp}{spp}{samples per pixel}
\newacronym{cnn}{CNN}{convolutional neural network}
\newacronym{gan}{GAN}{generative adversarial network}
\newacronym{mlp}{MLP}{multi-layer perceptron}
\newacronym{rmse}{rMSE}{relative mean squared error}
\newacronym{sbmc}{SBMC}{Sample-Based Monte Carlo Denoising}
\newacronym{gpu}{GPU}{graphics processing unit}
\newacronym{dssim}{DSSIM}{structural dissimilarity index}
\glsunset{dssim}  %

\copyrightyear{2025}
\acmYear{2025}
\setcopyright{cc}
\setcctype{by}
\acmConference[SA Conference Papers '25]{SIGGRAPH Asia 2025 Conference Papers}{December 15--18, 2025}{Hong Kong, Hong Kong}
\acmBooktitle{SIGGRAPH Asia 2025 Conference Papers (SA Conference Papers '25), December 15--18, 2025, Hong Kong, Hong Kong}
\acmDOI{10.1145/3757377.3763931}
\acmISBN{979-8-4007-2137-3/2025/12}

\begin{document}

\title{Nonlinear Noise2Noise for Efficient Monte Carlo Denoiser Training}

\author{Andrew Tinits}
\orcid{0000-0001-7593-3356}
\affiliation{%
  \institution{University of Waterloo}
  \city{Waterloo}
  \state{Ontario}
  \country{Canada}
}
\email{amtinits@uwaterloo.ca}

\author{Stephen Mann}
\orcid{0000-0001-8528-2921}
\affiliation{%
  \institution{University of Waterloo}
  \city{Waterloo}
  \state{Ontario}
  \country{Canada}
}
\email{smann@uwaterloo.ca}

\begin{abstract}
    The Noise2Noise method allows for training machine learning-based denoisers with pairs of input
    and target images where both the input and target can be noisy. This removes the need for
    training with clean target images, which can be difficult to obtain. However, Noise2Noise
    training has a major limitation: nonlinear functions applied to the noisy targets will skew the
    results. This bias occurs because the nonlinearity makes the expected value of the noisy targets
    different from the clean target image. Since nonlinear functions are common in image processing,
    avoiding them limits the types of preprocessing that can be performed on the noisy targets. Our
    main insight is that certain nonlinear functions can be applied to the noisy targets without
    adding significant bias to the results. We develop a theoretical framework for analyzing the
    effects of these nonlinearities, and describe a class of nonlinear functions with minimal bias.

    We demonstrate our method on the denoising of high dynamic range (HDR) images produced by Monte
    Carlo rendering, where generating high-sample count reference images can be prohibitively
    expensive. Noise2Noise training can have trouble with HDR images, where the training process is
    overwhelmed by outliers and performs poorly. We consider a commonly used method of addressing
    these training issues: applying a nonlinear tone mapping function to the model output and target
    images to reduce their dynamic range. This method was previously thought to be incompatible with
    Noise2Noise training because of the nonlinearities involved.  We show that certain combinations
    of loss functions and tone mapping functions can reduce the effect of outliers while introducing
    minimal bias. We apply our method to an existing machine learning-based Monte Carlo denoiser,
    where the original implementation was trained with high-sample count reference images. Our
    results approach those of the original implementation, but are produced using only noisy
    training data.
\end{abstract}



\begin{teaserfigure}
    \centering
    \includegraphics[width=\textwidth]{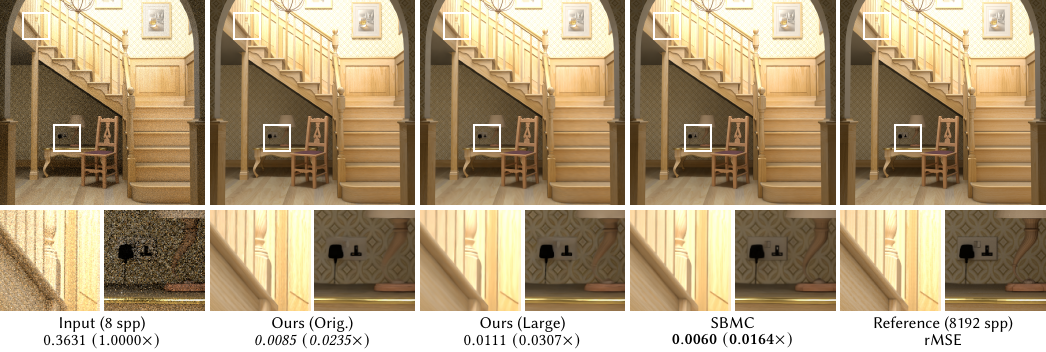}
    \caption{A Monte Carlo rendering at 8 samples per pixel (spp), denoised with our models,
    denoised with the SBMC model, and the reference image rendered at 8192~spp.  Our models are
    trained with only noisy data, while SBMC is trained with clean reference images.
    (\href{https://blendswap.com/blend/14449}{The Wooden Staircase} by
    \href{https://blendswap.com/profile/130393}{Wig42},
    \href{https://creativecommons.org/licenses/by/3.0/}{CC~BY~3.0})}
    \label{fig:comparison_full}
    \Description{A row of five images, each displaying a rendering of a wooden staircase inside a
    house. Below is a row of crops of the main images. The first image is noisy, while the remaining
    four images all look similar in quality, with no visible noise.}
\end{teaserfigure}

\maketitle

\section{Introduction}

There are many applications in image processing where clean images are hard to obtain, such as with
long-exposure photography or Monte Carlo rendering.  In these cases, the noise is gradually reduced
through a lengthy process, making clean images expensive to produce.  Denoising offers a solution to
this problem, where the image generating process can be stopped before convergence, and
post-processing techniques can be applied to reduce the remaining noise to an acceptable level.
Denoising has contributed to recent successes in adapting Monte Carlo rendering for real-time
applications, for example.  Common denoising methods use either hand-tuned filters or machine
learning models to reduce the noise.

Learning-based denoisers are trained on large datasets of images and their associated feature data.
In the supervised learning setting, these datasets contain pairs of noisy images with their
corresponding clean reference images.  However, difficulty in obtaining clean references presents a
challenge to training denoisers with supervised learning.  A potential solution to this problem is
to remove the need for clean reference images by changing to a different learning setting where
clean references are not required.  One such method is Noise2Noise, a weakly supervised learning
approach where denoisers are trained with pairs of noisy images~\citep{lehtinen_noise2noise_2018}.
For the common case where the noise distribution has zero mean, Noise2Noise is used with the $L_2$
loss to recover the expectation of the noisy targets $\E[\hat{y}]$, which is equal to the clean
target $y$~\citep{lehtinen_noise2noise_2018}.  The error from the noisy targets approaches zero as
the number of training examples increases~\citep{lehtinen_noise2noise_2018}.

Noise2Noise, in the zero-mean noise setting, has a major limitation: nonlinear functions applied to
the noisy targets will skew the results~\citep{lehtinen_noise2noise_2018}.  This bias occurs because
the nonlinearity $\varphi$ makes the expected value of the noisy targets $\E[\varphi(\hat{y})]$
different from the clean target $\varphi(\E[\hat{y}])$~\citep{lehtinen_noise2noise_2018}, since in
general $\E[\varphi(\hat{y})]\neq\varphi(\E[\hat{y}])$.  Nonlinear functions are common in image
processing, so the current practice of avoiding nonlinearity limits the types of preprocessing that
can be performed on the noisy targets.  Our main insight is that certain nonlinear functions can be
applied to the noisy targets without adding significant bias to the results.  We develop a
theoretical framework, based on the Jensen gap
$\E[\varphi(\hat{y})]-\varphi(\E[\hat{y}])$~\citep{jensen_sur_1906}, for analyzing the effects of
these nonlinearities on Noise2Noise training. We then use our framework to describe a class of
nonlinear functions with minimal bias.

Noise2Noise training has been widely adopted for other denoising problems, but has not seen wide
adoption for the denoising of Monte Carlo rendered images.  Monte Carlo rendering is a randomized
algorithm that samples the space of possible light paths through a scene.  Although this process
converges in the limit to the correct image, achieving a low noise level can require many thousands
of samples, which can be prohibitively expensive.  Monte Carlo noise has zero mean, which makes it
an ideal candidate for Noise2Noise training.  However, Monte Carlo renderings often have a \gls{hdr}
spanning many orders of magnitude, especially at low \gls{spp}.  Noise2Noise training can have
trouble with \gls{hdr} images, where the $L_2$ loss is overwhelmed by outliers and the training
fails to converge~\citep{lehtinen_noise2noise_2018}.  The Noise2Noise authors offer a solution to
this problem with their $L_\mathrm{HDR}$ loss function~\citep{lehtinen_noise2noise_2018}, but in our
testing this solution is insufficient, and the training still performs poorly.  We attribute this
difference in our findings to our more complex model and our significantly larger and more varied
training dataset.

One existing solution to the \gls{hdr} training issues is to apply a nonlinear tone mapping function
to the model output and target images to reduce their dynamic
range~\citep{gharbi_sample-based_2019}.  This method was previously thought to be incompatible with
Noise2Noise training because of the nonlinearities involved~\citep{lehtinen_noise2noise_2018}.  We
show that this method can in fact be used with Noise2Noise training.  We apply our theoretical
framework to show that certain combinations of loss functions and tone mapping functions can reduce
the effect of outliers while introducing minimal bias to the results.  Our experimental results
confirm the effectiveness of this approach.  We apply our method to an existing machine
learning-based Monte Carlo denoiser~\citep{gharbi_sample-based_2019}, where the original
implementation was trained with high-sample count reference images.  Our denoising results approach
those of the original implementation, but are produced using only low-sample count noisy training
data.  Monte Carlo denoising of varied scenes with complex models is therefore an example of an
application where training from noisy data is made possible by our nonlinear Noise2Noise method.

\section{Related Work}

\subsection{Weakly Supervised Image Denoising}

Noise2Noise~\citep{lehtinen_noise2noise_2018} was not the first work to recognize that clean
reference images are not necessary for image denoising.  Outside of machine learning methods, there
are classic algorithms such as NLM~\citep{buades_non-local_2005} and BM3D~\citep{dabov_image_2007}
that denoise a single noisy image without requiring any additional information.  These methods look
at statistical information from the rest of the image to determine how to denoise a particular
pixel.  Within machine learning, Noise2Noise is the first denoising method that can learn from noisy
images without requiring prior knowledge of either the noise distribution or the distribution of the
clean images~\citep{lehtinen_noise2noise_2018}.  Other methods that can train on noisy images, such
as AmbientGAN~\citep{bora_ambientgan_2018} and GradNet~\citep{guo_gradnet_2019}, require either an
explicit statistical model of the noise (AmbientGAN), or additional regularization that makes
assumptions about the clean results (GradNet).  In the case of Monte Carlo denoising, the noise
distribution cannot be characterized analytically, so methods with the former requirement cannot be
used~\citep{lehtinen_noise2noise_2018}.

Since the publication of Noise2Noise, several works have extended the idea so that only individual
noisy images are required for training, rather than the noisy image pairs required by Noise2Noise
\citep{lehtinen_noise2noise_2018}.  These works include Noise2Void~\citep{krull_noise2void_2019},
Noise2Self~\citep{batson_noise2self_2019}, and the work of \citet{laine_high-quality_2019}.  Our
idea of applying nonlinearity to the noisy targets can equally be used with these methods and others
that make use of the Noise2Noise property.  Single-image denoising is useful in settings where noisy
image pairs are difficult to obtain, such as in biomedical imaging. However, these methods generally
come with reduced performance when compared to Noise2Noise because they have less information
available during training~\citep{krull_noise2void_2019}.  It therefore makes sense to use
Noise2Noise in settings such as Monte Carlo denoising, where noisy image pairs are easy to obtain.

This work is a continuation of previous experimental work by the first
author~\citep{tinits_learning_2022}.  To our knowledge, these two works are the
first to successfully apply nonlinear functions to the noisy targets in Noise2Noise training.

\subsection{Monte Carlo Denoising}

Monte Carlo denoising methods can be separated into two main categories: traditional algorithms, and
machine learning-based methods.  We briefly review machine-learning based methods below.  For a more
comprehensive review, see the survey papers by \citet{zwicker_recent_2015} and
\citet{huo_survey_2021}.

The use of machine learning for Monte Carlo denoising was initiated by
\citet{kalantari_machine_2015}, who used a \gls{mlp} to predict the parameters of a cross-bilateral
filter.  \citeauthor{bako_kernel-predicting_2017} improved this idea by using a deep \gls{cnn} to
predict filter kernels separately for each pixel~\citeyearpar{bako_kernel-predicting_2017}.  This
kernel prediction idea is used by many later
works~\citep{vogels_denoising_2018,gharbi_sample-based_2019,munkberg_neural_2020,back_deep_2020,lin_detail_2020,meng_real-time_2020,lin_path-based_2021}.
Other recent works take the simpler approach of predicting the pixel radiances
directly~\citep{chaitanya_interactive_2017,xu_adversarial_2019,yang_fast_2018,yang_demc_2019,wong_deep_2019,guo_gradnet_2019,lu_dmcr-gan_2020}.
\citeauthor{gharbi_sample-based_2019} propose another alternative where splatting kernels are used
to allow individual pixels to determine their own
contributions~\citeyearpar{gharbi_sample-based_2019}.  This concept is extended by
\citet{munkberg_neural_2020}, who divide the samples into layers and filter them separately to
improve efficiency, and by \citet{lin_path-based_2021} and \citet{cho_weakly-supervised_2021}, who
add information about individual Monte Carlo paths.  Other recent works use different model
architectures, such as \glspl{gan}~\citep{lu_denoising_2021,lu_dmcr-gan_2020,xu_adversarial_2019},
self-attention~\citep{chen_temporally_2024,oh_joint_2024,chen_monte_2023,yu_monte_2021}, and
diffusion models~\citep{vavilala_denoising_2024}.  Several works in recent years have explored
unsupervised
learning~\citep{choi_online_2024,back_self-supervised_2022,xu_unsupervised_2020,guo_gradnet_2019}.
Weakly supervised learning for Monte Carlo denoising was studied by
\citet{cho_weakly-supervised_2021}, but their method uses high-sample count reference images for
training.

\section{Noise2Noise Theory}
\label{sec:n2n_theory}

The typical supervised machine learning regression task uses a model $f$ with parameters $\theta$, a
loss function $L$, and training data composed of pairs of example inputs and targets $(x_i,y_i)$.
Representing the inputs and targets with random variables $x$ and $y$, the training procedure finds
the parameters that result in the lowest average loss over the training data:
\begin{equation}\label{eq:1}
    \argmin_\theta \E_{x,y}[L(f_\theta(x),y)]\,.
\end{equation}

As noted by the Noise2Noise authors, this training procedure is equivalent to performing the
minimization separately for each input~\citep{lehtinen_noise2noise_2018}, so expression~\eqref{eq:1}
becomes
\begin{equation}\label{eq:2}
    \argmin_\theta \E_x\left[\E_{y\mid x}[L(f_\theta(x),y)]\right]\,.
\end{equation}

The Noise2Noise authors then show that the estimate is the same if the training targets $y$ are
replaced with any random variable $\hat{y}$ whose expected value
$\E[\hat{y}]=y$~\citep{lehtinen_noise2noise_2018}.  In the denoising scenario, we can therefore
replace the clean training targets with noisy ones without changing the result, as long as the noise
has zero mean~\citep{lehtinen_noise2noise_2018}.  For denoising, the training inputs are also noisy
estimates $\hat{x}$ of the clean targets $y$ (not necessarily from the same distribution as the
noisy targets $\hat{y}$), such that $\E[\hat{y}\mid\hat{x}]=y$~\citep{lehtinen_noise2noise_2018}.
Substituting $\hat{x}$ and $\hat{y}$ into expression~\eqref{eq:2} results in the Noise2Noise
training procedure:
\begin{equation}\label{eq:3}
    \argmin_\theta \E_{\hat{x}}\left[\E_{\hat{y}\mid\hat{x}}[L(f_\theta(\hat{x}),\hat{y})]\right]\,.
\end{equation}

\section{Nonlinear Noise2Noise}
\label{sec:nonlinear}

\subsection{Loss Functions}
\label{ssec:loss_functions}

Since the choice of loss function affects the training results, we start by analyzing the effects of
some relevant loss functions on the Noise2Noise training process.  As shown in
Section~\ref{sec:n2n_theory}, the target of the training is the minimum of the expected loss, which
can be minimized separately for each input $\hat{x}$.  Optimizing the model parameters $\theta$ is
equivalent to optimizing the model output $f_\theta(\hat{x})$.  So, defining
$\tilde{y}=f_\theta(\hat{x})$, expression~\eqref{eq:3} can therefore be minimized by finding the
value of $\tilde{y}$ that minimizes the inner expectation:
\begin{equation}\label{eq:4}
    \argmin_{\tilde{y}} \E_{\hat{y}\mid\hat{x}}[L(\tilde{y},\hat{y})]\,.
\end{equation}

For some loss functions, the value of $\tilde{y}$ that minimizes expression~\eqref{eq:4} can be
found analytically.  We first derive the well-known result that the $L_2$ loss,
$L_2(\tilde{y},\hat{y}) = (\tilde{y}-\hat{y})^2$, has its minimum at the mean, which in our case is
the clean target $y$.  This result can be found by setting the partial derivative to zero, and
solving for $\tilde{y}$:
\begin{align*}
    0 &= \frac{\partial}{\partial\tilde{y}}\E_{\hat{y}\mid\hat{x}}\left[(\tilde{y}-\hat{y})^2\right] \\
      &= \frac{\partial}{\partial\tilde{y}}\left(\tilde{y}^2-2\tilde{y}\E_{\hat{y}\mid\hat{x}}[\hat{y}]+\E_{\hat{y}\mid\hat{x}}[\hat{y}^2]\right) \\
      &= 2\tilde{y}-2\E_{\hat{y}\mid\hat{x}}[\hat{y}] \\
    \tilde{y} &= \E_{\hat{y}\mid\hat{x}}[\hat{y}] = y\,.
\end{align*}
We can verify that this critical point is a minimum by looking at the second partial derivative,
$\frac{\partial^2}{\partial^2\tilde{y}}\E_{\hat{y}\mid\hat{x}}\left[(\tilde{y}-\hat{y})^2\right]=2$.
This value is strictly greater than zero for all values of $\tilde{y}$, so the $L_2$ loss is
strictly convex, and $y$ is its single global minimum.

We next derive the minimum for the \gls{rmse} loss, $L_\mathrm{rMSE} =
(\tilde{y}-\hat{y})^2/(\hat{y}+\epsilon)^2$, which is designed to minimize the effect of outliers in
\gls{hdr} images~\citep{rousselle_adaptive_2011}.  We again set the partial derivative to zero and
solve for $\tilde{y}$:
\begin{align}
    0 &= \frac{\partial}{\partial\tilde{y}}\E_{\hat{y}\mid\hat{x}}\left[\frac{(\tilde{y}-\hat{y})^2}{(\hat{y}+\epsilon)^2}\right] \nonumber\\
      &= 2\tilde{y}\E_{\hat{y}\mid\hat{x}}\left[\frac{1}{(\hat{y}+\epsilon)^2}\right]-2\E_{\hat{y}\mid\hat{x}}\left[\frac{\hat{y}}{(\hat{y}+\epsilon)^2}\right] \nonumber\\
    \tilde{y} &= \frac{\E_{\hat{y}\mid\hat{x}}\left[\frac{\hat{y}}{(\hat{y}+\epsilon)^2}\right]}{\E_{\hat{y}\mid\hat{x}}\left[\frac{1}{(\hat{y}+\epsilon)^2}\right]}
    \approx \frac{\E_{\hat{y}\mid\hat{x}}\left[\frac{1}{\hat{y}}\right]}{\E_{\hat{y}\mid\hat{x}}\left[\frac{1}{\hat{y}^2}\right]}\,. \label{eq:5}
\end{align}
This value is not easy to simplify because of the nonlinear functions involved, since in general
$\E[\varphi(\hat{y})]\neq\varphi(\E[\hat{y}])$.  We will show how to analyze this value further in
Section~\ref{ssec:jensen_bound}.  For now, we can verify that this critical point is a minimum by
looking at the first partial derivative above.  Our variables represent images and are therefore
nonnegative (ignoring negatives from color space transformations or negative filter lobes), so the
derivative will be positive when
\begin{align*}
    2\tilde{y}\E_{\hat{y}\mid\hat{x}}\left[\frac{1}{(\hat{y}+\epsilon)^2}\right]&>2\E_{\hat{y}\mid\hat{x}}\left[\frac{\hat{y}}{(\hat{y}+\epsilon)^2}\right] \\
    \tilde{y} &> \frac{\E_{\hat{y}\mid\hat{x}}\left[\frac{\hat{y}}{(\hat{y}+\epsilon)^2}\right]}{\E_{\hat{y}\mid\hat{x}}\left[\frac{1}{(\hat{y}+\epsilon)^2}\right]}\,,
\end{align*}
which is the value in equation~\eqref{eq:5}.  Similarly, the derivative will be zero at this value,
and negative when $\tilde{y}$ is less than this value.  $L_\mathrm{rMSE}$ is therefore pseudoconvex
for $\tilde{y}\geq0$, and so the local minimum in equation~\eqref{eq:5} is also the single global
minimum~\citep{mangasarian_pseudo-convex_1965}.

The final loss function we consider is the $L_\mathrm{HDR}$ loss, $L_\mathrm{HDR} =
(\tilde{y}-\hat{y})^2/(\tilde{y}+\epsilon)^2$, which is proposed by the Noise2Noise authors as an
alternative to $L_\mathrm{rMSE}$ that avoids the nonlinearity
issues~\citep{lehtinen_noise2noise_2018}.  Solving the partial derivative again, we have
\begin{align}
    0 &= \frac{\partial}{\partial\tilde{y}}\E_{\hat{y}\mid\hat{x}}\left[\frac{(\tilde{y}-\hat{y})^2}{(\tilde{y}+\epsilon)^2}\right] \nonumber\\
      &= \frac{2\tilde{y}}{(\tilde{y}+\epsilon)^3}\left(\E_{\hat{y}\mid\hat{x}}[\hat{y}]+\epsilon\right)-\frac{2}{(\tilde{y}+\epsilon)^3}\left(\E_{\hat{y}\mid\hat{x}}[\hat{y}^2]+\epsilon\E_{\hat{y}\mid\hat{x}}[\hat{y}]\right) \nonumber\\
    \tilde{y} &= \frac{\E_{\hat{y}\mid\hat{x}}[\hat{y}^2]+\epsilon\E_{\hat{y}\mid\hat{x}}[\hat{y}]}{\E_{\hat{y}\mid\hat{x}}[\hat{y}]+\epsilon}
         \approx \frac{\E_{\hat{y}\mid\hat{x}}[\hat{y}^2]}{\E_{\hat{y}\mid\hat{x}}[\hat{y}]}
               = y + \frac{\Var(\hat{y})}{y}\,, \label{eq:6}
\end{align}
where the last equality comes from the definition of variance: $\Var(X) = \E[X^2]-\E[X]^2$.  The
nonlinearity is easily removed in this case, revealing a bias term of the variance divided by the
mean, which is also known as the index of dispersion.  We can see that, in the same way as
$L_\mathrm{rMSE}$, the first partial derivative above is negative before the minimum at
equation~\eqref{eq:6}, zero at the minimum, and positive after.  $L_\mathrm{HDR}$ is therefore
pseudoconvex for $\tilde{y}\geq0$, and so the local minimum in equation~\eqref{eq:6} is also the
single global minimum~\citep{mangasarian_pseudo-convex_1965}.  Since the denominator can be
separated from the expectation, $L_\mathrm{HDR}$ can also be defined with the gradient of the
denominator set to zero, which removes the bias~\citep{lehtinen_noise2noise_2018}.  We call this
definition $L_\mathrm{HDR}^\ast$, and we evaluate it separately in
Section~\ref{ssec:model_selection}.

\subsection{Loss Functions with Nonlinearity}
\label{ssec:nonlinearity}

We now introduce our method, where a nonlinear function $T(v)$ is applied to the model output
$\tilde{y}$ and the noisy target $\hat{y}$ before they are passed to the loss function $L$, which
then becomes $L(T(\tilde{y}),T(\hat{y}))$.  $T(v)$ must be twice differentiable, strictly monotonic,
and nonnegative for $v\geq0$.  In our example application, $T(v)$ is a tone mapping function that
reduces the dynamic range of the loss function inputs.

We now analyze the same loss functions as in Section~\ref{ssec:loss_functions}, but with $T(v)$
applied to their inputs.  Starting with the $L_2$ loss, we have:
\begin{align}
    0 &= \frac{\partial}{\partial\tilde{y}}\E_{\hat{y}\mid\hat{x}}\left[(T(\tilde{y})-T(\hat{y}))^2\right] \nonumber\\
      &= \frac{\partial}{\partial\tilde{y}}\left(T(\tilde{y})^2-2T(\tilde{y})\E_{\hat{y}\mid\hat{x}}[T(\hat{y})]+\E_{\hat{y}\mid\hat{x}}[T(\hat{y})^2]\right) \nonumber\\
      &= 2T(\tilde{y})T'(\tilde{y})-2T'(\tilde{y})\E_{\hat{y}\mid\hat{x}}[T(\hat{y})] \nonumber\\
    \tilde{y} &= T^{-1}\left(\E_{\hat{y}\mid\hat{x}}[T(\hat{y})]\right)\,. \label{eq:7}
\end{align}
To show that this value is a minimum, we look at the first partial derivative above, which is
positive when
\begin{align*}
    2T(\tilde{y})T'(\tilde{y}) &> 2T'(\tilde{y})\E_{\hat{y}\mid\hat{x}}[T(\hat{y})] \\
    \tilde{y} &> T^{-1}\left(\E_{\hat{y}\mid\hat{x}}[T(\hat{y})]\right)\,,
\end{align*}
zero at this value, and negative otherwise.  For the last line above, the direction of the
inequality changes twice if $T(v)$ is strictly decreasing, resulting in no change to the direction.
The expected loss is therefore pseudoconvex for $\tilde{y}\geq0$, and so the local minimum in
equation~\eqref{eq:7} is also the single global minimum~\citep{mangasarian_pseudo-convex_1965}.

The $L_\mathrm{rMSE}$ loss becomes $(T(\tilde{y})-T(\hat{y}))^2/(T(\hat{y})+\epsilon)^2$ with our
method, which has its single global minimum at
\begin{equation}\label{eq:8}
    \tilde{y} = T^{-1}\left(\frac{\E_{\hat{y}\mid\hat{x}}\left[\frac{T(\hat{y})}{(T(\hat{y})+\epsilon)^2}\right]}{\E_{\hat{y}\mid\hat{x}}\left[\frac{1}{(T(\hat{y})+\epsilon)^2}\right]}\right)
        \approx T^{-1}\left(\frac{\E_{\hat{y}\mid\hat{x}}\left[\frac{1}{T(\hat{y})}\right]}{\E_{\hat{y}\mid\hat{x}}\left[\frac{1}{T(\hat{y})^2}\right]}\right)\,,
\end{equation}
while the $L_\mathrm{HDR}$ loss becomes $(T(\tilde{y})-T(\hat{y}))^2/(T(\tilde{y})+\epsilon)^2$ with
our method, which has its single global minimum at
\begin{equation}\label{eq:9}
    \tilde{y} = T^{-1}\left(\frac{\E_{\hat{y}\mid\hat{x}}[T(\hat{y})^2]+\epsilon\E_{\hat{y}\mid\hat{x}}[T(\hat{y})]}{\E_{\hat{y}\mid\hat{x}}[T(\hat{y})]+\epsilon}\right)
        \approx T^{-1}\left(\frac{\E_{\hat{y}\mid\hat{x}}[T(\hat{y})^2]}{\E_{\hat{y}\mid\hat{x}}[T(\hat{y})]}\right)\,,
\end{equation}
using the same methods as above.

The meaning of these minimum values is not immediately obvious because of the nonlinear functions
present inside of the expectations.  In the next section, we will show how to derive bounds on the
bias introduced by these nonlinearities, and in the following section we will show how to ensure
that these bounds are as small as possible.

\subsection{Bounding the Jensen Gap}
\label{ssec:jensen_bound}

Jensen's inequality states that $\E[\varphi(X)]\geq\varphi(\E[X])$ for a finite random variable $X$
when $\varphi$ is a convex function~\citep{jensen_sur_1906}.  The difference between the two sides
of the inequality, $\E[\varphi(X)]-\varphi(\E[X])$, is known as the Jensen gap.  We use the method
of \citet{liao_sharpening_2019} to derive bounds for the Jensen gap, which only requires that
$\varphi$ is twice differentiable (not necessarily convex).  With this method, lower and upper
bounds for the Jensen gap can be derived with the form $J(\mu)\Var(X)$, where $J$ is a function of
the mean $\mu$ that is related to the curvature of the nonlinear function $\varphi$.  Intuitively,
the Jensen gap bound will be smaller when $\varphi$ is closer to being linear, especially around the
mean $\mu$.  The bound will also be smaller when the variance $\Var(X)$ is smaller, since the random
samples will be exposed to less of the nonlinearity of $\varphi$.  In particular, the Jensen gap
bound of \citet{liao_sharpening_2019} is
\begin{equation}\label{eq:10}
    J_-(\mu)\Var(X) \leq \E[\varphi(X)] - \varphi(\E[X]) \leq J_+(\mu)\Var(X)\,,
\end{equation}
where
\begin{equation*}
    J_-(\mu)=\inf_{x\in(a,b)}h(x,\mu) \enspace\text{and}\enspace J_+(\mu)=\sup_{x\in(a,b)}h(x,\mu)\,,
\end{equation*}
and
\begin{equation*}
    h(x,\mu) = \frac{\varphi(x)-\varphi(\mu)}{(x-\mu)^2} - \frac{\varphi'(\mu)}{x-\mu}\,.
\end{equation*}
In our case, we have $a=0$, $b=\infty$, $X=\hat{y}$, and $\mu=y$.  Note that the variance identity
applied in equation~\eqref{eq:6} is a special case of inequality~\eqref{eq:10}, where $J_-=J_+=1$
for $\varphi(X)=X^2$.  Note as well that it is possible to derive tighter bounds on the Jensen gap
using higher central moments than $\Var(X)$~\citep{lee_further_2021}, but these moments are less
interpretable for arbitrary distributions.

\begin{figure*}[t]
    \centering
    \includegraphics[width=\textwidth]{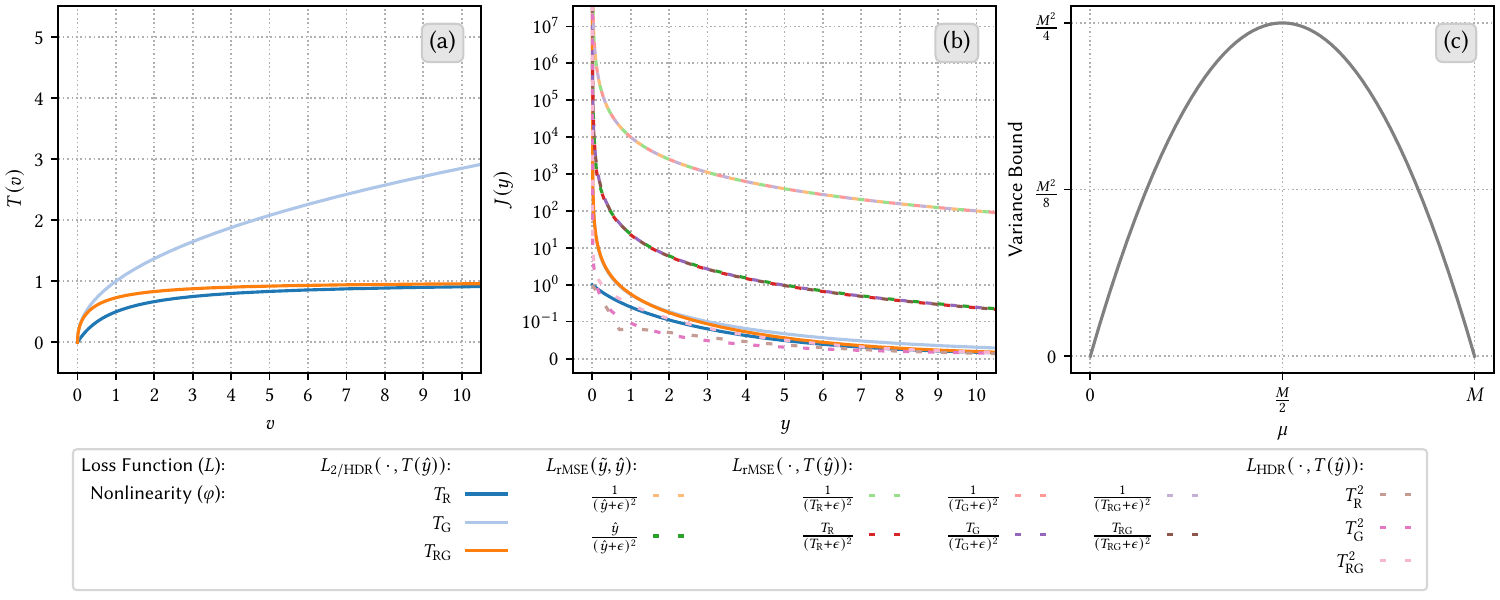}
    \caption{(a) Tone mapping functions used for our experimental results.  See
    Section~\ref{ssec:model_selection} for details.  (b)  $J(y)$ functions for the listed
    $\varphi(\hat{y})$ functions resulting from all combinations of our studied loss functions and
    tone mapping functions.  $J_-(y)$ and $J_+(y)$ functions are represented together as
    $\max(\abs{J_-(y)}\,,\abs{J_+(y)})$, with $\epsilon=0.01$.  See Section~\ref{ssec:jensen_bound}
    for details on $J(y)$ functions, and Section~\ref{ssec:jensen_gap} for a list of the $J(y)$ functions plotted
    above.  (c) Graphical representation of the Bhatia-Davis inequality for bounded probability
    distributions with minimum $m=0$ and maximum $M$.  See Section~\ref{ssec:minimizing} for
    details.}
    \label{fig:curves}
    \Description{Three plots, labeled (a), (b), and (c).  Plot (a) shows three curves starting at
    the origin and increasing to the right.  The T-R and T-RG curves are bounded at 1, while the T-G
    curve is unbounded.  Plot (b) shows 14 curves, all apparently exponentially decreasing to the
    right.  Most of the curves are infinite at y=0, while the T-R and T-R squared curves have a
    J(y)=1 value at y=0.  Plot (c) shows a parabola stretching from mu=0 to mu=M.  The maximum value
    of one quarter M squared occurs at a mu value of one half M.}
\end{figure*}

We now apply the method of \citet{liao_sharpening_2019} to the minimum expected loss values
derived in Section~\ref{ssec:nonlinearity}.  Starting with the $L_2(T(\tilde{y}),T(\hat{y}))$ loss,
the nonlinearity in equation~\eqref{eq:7} is $T(\hat{y})$.  We therefore set $\varphi(X)=T(\hat{y})$
in inequality~\eqref{eq:10}, giving
\begin{gather*}
    J_-(y)\Var(\hat{y}) \leq \E[T(\hat{y})] - T(\E[\hat{y}]) \leq J_+(y)\Var(\hat{y}) \\
    T(y) + J_-(y)\Var(\hat{y}) \leq \E[T(\hat{y})] \leq T(y) + J_+(y)\Var(\hat{y}) \\
    T^{-1}(T(y) + J_-(y)\Var(\hat{y})) \leq \tilde{y} \leq T^{-1}(T(y) + J_+(y)\Var(\hat{y}))\,,
\end{gather*}
where in the last line we have applied equation~\eqref{eq:7}.  Note that both inequalities change
direction in the last line if $T(v)$ is strictly decreasing.  Intuitively, we have bias terms
of the form $J(y)\Var(\hat{y})$ that are applied in the nonlinear space defined by $T(v)$, and then
the results are mapped back to the linear space by $T^{-1}(v)$.  We assume that the $J_-$ terms do
not make the resulting values negative.

For the $L_\mathrm{rMSE}(T(\tilde{y}),T(\hat{y}))$ loss, we apply the same procedure as above to
equation~\eqref{eq:8} to obtain an upper bound of
\begin{equation*}
    \tilde{y} \leq T^{-1}\left(\frac{\frac{T(y)}{(T(y)+\epsilon)^2}+J_+^{-1}(y)\Var(\hat{y})}{\frac{1}{(T(y)+\epsilon)^2}+J_-^{-2}(y)\Var(\hat{y})}\right) \,,
\end{equation*}
where again the inequality changes direction if $T(v)$ is strictly decreasing (making this the lower
bound).  To obtain the other bound, change the direction of the inequality, and swap the subscripts
of the $J(y)$ functions.  Here, $J_{+/-}^{-1}(y)$ and $J_{+/-}^{-2}(y)$ mean the $J_{+/-}(y)$
functions for $\varphi(\hat{y})\!=\!\frac{T(\hat{y})}{(T(\hat{y})+\epsilon)^2}$ and
$\varphi(\hat{y})\!=\!\frac{1}{(T(\hat{y})+\epsilon)^2}$, respectively.

For the $L_\mathrm{HDR}(T(\tilde{y}),T(\hat{y}))$ loss, we apply the same procedure to
equation~\eqref{eq:9} to obtain an upper bound of
\begin{equation*}
    \tilde{y} \leq T^{-1}\left(\frac{T(y)^2+J_+^2(y)\Var(\hat{y})+\epsilon(T(y)+J_+(y)\Var(\hat{y}))}{T(y)+J_-(y)\Var(\hat{y})+\epsilon}\right) \,,
\end{equation*}
where again the inequality changes direction if $T(v)$ is strictly decreasing (making this the lower
bound).  To obtain the other bound, change the direction of the inequality, and swap the subscripts
of the $J(y)$ functions.  Here, $J_{+/-}^2(y)$ and $J_{+/-}(y)$ mean the $J_{+/-}(y)$ functions for
$\varphi(\hat{y})=T(\hat{y})^2$ and $\varphi(\hat{y})=T(\hat{y})$, respectively.

We can also now derive the Jensen gap bound for the original $L_\mathrm{rMSE}(\tilde{y},\hat{y})$
loss, as promised in Section~\ref{ssec:loss_functions}.  Applying the same procedure as above to
equation~\eqref{eq:5}, we obtain a bound of
\begin{equation*}
    \frac{\frac{y}{(y+\epsilon)^2}+J_-^{-1}(y)\Var(\hat{y})}{\frac{1}{(y+\epsilon)^2}+J_+^{-2}(y)\Var(\hat{y})} \leq
    \tilde{y} \leq
    \frac{\frac{y}{(y+\epsilon)^2}+J_+^{-1}(y)\Var(\hat{y})}{\frac{1}{(y+\epsilon)^2}+J_-^{-2}(y)\Var(\hat{y})} \,,
\end{equation*}
where $J_{+/-}^{-1}(y)$ and $J_{+/-}^{-2}(y)$ mean the $J_{+/-}(y)$ functions for
$\varphi(\hat{y})\!=\!\frac{\hat{y}}{(\hat{y}+\epsilon)^2}$ and
$\varphi(\hat{y})\!=\!\frac{1}{(\hat{y}+\epsilon)^2}$, respectively.

The $J(y)$ functions for the tone mapping functions we evaluate in our experiments (see
Section~\ref{ssec:model_selection}) are plotted in Figure~\ref{fig:curves}(b), and are listed in
Section~\ref{ssec:jensen_gap}.

\subsection{Minimizing the Jensen Gap}
\label{ssec:minimizing}

In Section~\ref{ssec:jensen_bound} we showed how to derive Jensen gap bounds for nonlinear functions
applied to some relevant loss functions.  We now show how to ensure that these bounds are as small
as possible.  There are two cases where the bounds derived above are small, namely when
$\Var(\hat{y})$ is small, or when $\varphi(\hat{y})$ is nearly linear~\citep{liao_sharpening_2019}.
The first case, where $\Var(\hat{y})$ is small, can be difficult to achieve in practice.  For
applications like long-exposure photography or Monte Carlo rendering, the noise is gradually reduced
over time, and zero noise is only reached after an infinite amount of time.  Since we only have a
finite amount of time in practice, there will always be some amount of noise remaining, even in the
so-called ``clean'' target images.  So, in a sense, we are always doing Noise2Noise training in
these situations, even when training with ``clean'' targets~\citep{tinits_learning_2022}.
Therefore, the Jensen gap bound is a concern even with ``clean'' targets, although the effect will
be larger when the targets have a significant amount of noise.

The other case where the Jensen gap is small is when $\varphi(\hat{y})$ is nearly linear.  This is
acceptable when nonlinear functions are not needed, but there are practical applications where
nonlinear functions are useful.  Dynamic range reduction, for example, is not possible with a linear
function, yet significantly improves the stability of the training process.  It is therefore
advisable to choose nonlinear functions that both have practical benefits, and also have low Jensen
gap bounds.  For example, the tone mapping functions that we use for our experimental results are
plotted in Figure~\ref{fig:curves}(a).  For effective dynamic range reduction, the function should
have more curvature near zero, and less towards infinity.  This property also results in low $J(y)$
values for most of the function domain, except near zero (see the solid lines in
Figure~\ref{fig:curves}(b)).  \Gls{hdr} images will therefore often have low $J(y)$ values for these
tone mapping functions, since the clean targets $y$ will often have large values.

There is also a separate reason for why the Jensen gap bound will be small when the clean target $y$
is small.  For most practical purposes, $y$ will in fact have a maximum possible value $M$.  This
value could come from the maximum representable value of the data format, the maximum possible
output value of the camera sensor, or the maximum brightness of the light sources in a rendered
scene, for example.  Given this maximum value $M$, the Bhatia-Davis inequality lets us bound the
variance $\Var(\hat{y})$ to
\begin{equation*}
    \Var(\hat{y}) \leq (M - y)(y - m) = -y^2 + My \,,
\end{equation*}
for the minimum value $m=0$~\citep{bhatia_better_2000}.  This bound forms a parabola with roots at
$y=0$ and $y=M$, and a maximum value of $\frac{M^2}{4}$ at $y=\frac{M}{2}$ (see
Figure~\ref{fig:curves}(c)).  This maximum value is also known as Popoviciu's
inequality~\citep{popoviciu_sur_1935}.  Because of the roots of the parabola, this bound on the
variance will be small for $y$ values near $y=0$ and $y=M$.

In summary, for a low Jensen gap bound, choose a nonlinear function with low curvature, especially
near where the clean targets will be located.  For bounded probability distributions, the function
can have some curvature near the minimum and maximum possible values.  In the next section, we
present our experimental results, and show that the practical benefits of certain nonlinear
functions can vastly outweigh the bias that they introduce to the results.

\section{Experimental Results}
\label{sec:results}

We demonstrate the effectiveness of our method by applying it to a Monte Carlo denoising task.  We
first describe our experimental methods (Section~\ref{ssec:methods}), then we evaluate several
combinations of loss functions and tone mapping functions (Section~\ref{ssec:model_selection}), and
then we compare our method to training with clean references (Section~\ref{ssec:comparison}).

\subsection{Methods}
\label{ssec:methods}

We apply our training method to an offline Monte Carlo denoising algorithm called
\gls{sbmc}~\citep{gharbi_sample-based_2019}.  We chose \gls{sbmc} as our denoiser because of its
adaptability to various simple loss functions, and because of its publicly available test set for
comparisons~\citep{gharbi_sample-based_2019}.  Our aim was to reproduce the denoising results of the
\gls{sbmc} authors, but using Noise2Noise training instead of the clean reference images used by the
original implementation.  This process involved generating randomized scenes for training, rendering
these scenes to noisy image pairs with additional feature data, training \gls{sbmc} models for
various combinations of loss functions and tone mapping functions, and evaluating their denoising
performance.

\paragraph{Scene generation}
The training scenes are randomly generated to include a wide variety of objects, materials, and
light transport scenarios.  This variety ensures that the resulting denoising models can handle a
diverse range of scenes.  We used the scene generator from
\gls{sbmc}~\citep{gharbi_sample-based_2019}, which generates both indoor and outdoor scenes.  For
indoor scenes, it uses a dataset of room models called SunCG~\citep{song_semantic_2017}, which is no
longer available~\citep{robitzski_startup_2019}, so our dataset consists of only outdoor scenes.  We
did not substitute another dataset for indoor scenes to avoid any factors that could artificially
improve our performance relative to \gls{sbmc}.  This difference could affect the ability of our
models to generalize to indoor scenes, and should be taken into account when comparing the
performance of our models to \gls{sbmc}.

\paragraph{Rendering}
We used the PBRT~v2 renderer~\citep{pharr_physically_2023} with the patch provided by \gls{sbmc} to
save the Monte Carlo samples separately and to gather additional feature
data~\citep{gharbi_sample-based_2019}.  We implemented Noise2Noise training by rendering two
different noisy images for each scene, each with different random seeds. One pair of noisy images is
rendered for each generated scene to maximize the diversity of the training dataset.

\paragraph{Dataset}
Our goal in generating a training dataset was to match the dataset used by \gls{sbmc}, to allow for
testing our method without changing any other variables.  To match \gls{sbmc}, we generated a
training dataset of around 300,000 example pairs at $128\times128$
resolution~\citep{gharbi_sample-based_2019}.  The exact number was 283,831 example pairs due to the
rejection sampling of the \gls{sbmc} scene generator.  For \gls{sbmc}, the noisy inputs were
rendered at 8~\gls{spp} and the references were rendered at
4096~\gls{spp}~\citep{gharbi_sample-based_2019}.  For our training set, the noisy images were both
rendered at 8~\gls{spp}.  We also generated a second training set by repeating this process with a
different starting seed, resulting in an additional 284,016 example pairs.  To monitor progress
during training, we also generated a validation set.  To match \gls{sbmc}, our validation set
contained around 1000 example pairs with references rendered at
4096~\gls{spp}~\citep{gharbi_sample-based_2019}, resulting in 937 example pairs after rejection
sampling.

\paragraph{Training}
We trained our models in the same way as \gls{sbmc}, with one small modification to improve
performance.  As with \gls{sbmc}, the number of input \gls{spp} was selected randomly from $2$ to
$8$ during training as a form of data augmentation, so that the models can handle different numbers
of input \gls{spp}~\citep{gharbi_sample-based_2019}.  With \gls{sbmc} this choice is made per
training example, while in our implementation the choice is made per training batch.  This change
allows for greater parallelism, and therefore improved training performance.  \gls{sbmc} has a fixed
batch size of one because of this limitation, whereas with our implementation the batch size can be
larger and split over multiple \glspl{gpu}.  Convergence should not be affected since the model is
still exposed to a variety of \gls{spp}.  In particular, we use a batch size of $16$ split over four
Nvidia Tesla T4 \glspl{gpu} with 16~GB of memory each.

\begin{figure*}[t]
    \centering
    \includegraphics[width=\textwidth]{all_three.pdf}
    \caption{(a)~Training losses, (b)~Gradient norms, and (c)~Validation \acrshort{rmse} for all
    combinations of loss functions, tone mapping functions, and tone mapping placement.  For the
    training losses, the units differ for the different loss functions, so comparisons in terms of
    loss value are not meaningful, and instead we are interested in the numerical stability of the
    loss landscapes.  The training losses and gradient norms are smoothed with a centered median
    filter with a width of 2,000 iterations.  The gradient norms were clipped at a value of $10^3$
    during training, and so will not exceed this value.  Training runs that end before 40,000
    iterations were terminated due to numerical errors.}
    \label{fig:tone_loss}
    \Description{Three plots, labeled (a), (b), and (c), containing training loss curves, gradient
    norm curves, and validation relative MSE curves, respectively.  All three plots have iterations
    on the X axis.  In the (a) plot, the curves with lower values are relatively smooth and flat,
    while the curves with higher values are more noisy.  In the (b) plot, most curves have a
    moderate amount of noise, while some curves shoot up to 1000 or drop down to 0 very quickly.  In
    the (c) plot, again the curves with lower values are relatively smooth and flat, while the
    curves with higher values are more noisy.}
\end{figure*}

\subsection{Model Selection}
\label{ssec:model_selection}

For some applications of nonlinear Noise2Noise, it might be necessary to apply our theoretical
analysis to design new nonlinear functions.  However, this is not needed for Monte Carlo denoising,
since certain combinations of existing loss functions and tone mapping functions meet our
theoretical criteria and perform well in practice.  Note that there is no universally optimal
combination of loss function and tone mapping function, because the performance depends on the
denoising model and data distribution.  Therefore, we choose the combination with the best
experimental performance on our model and dataset.

We consider the loss functions introduced in Section~\ref{ssec:loss_functions}, which correspond to
the losses used by \gls{sbmc} and
Noise2Noise~\citep{lehtinen_noise2noise_2018,gharbi_sample-based_2019}.  The plain $L_2$ loss
without tone mapping is unbiased, but is easily affected by outliers, so we do not expect it to
perform well in practice.  The $L_\mathrm{rMSE}$ and $L_\mathrm{HDR}$ losses are normalized to
reduce the impact of outliers.  However, our theoretical results show that the plain
$L_\mathrm{rMSE}$ and $L_\mathrm{HDR}$ losses have relatively high Jensen gap bounds (see
Figure~\ref{fig:curves}(b); $L_\mathrm{HDR}$ has a constant $J(y)=1$), so we do not expect strong
performance from these losses.  $L_\mathrm{HDR}^\ast$, on the other hand, is both normalized and
unbiased, and could therefore perform well if the normalization is effective.

We consider several common tone mapping functions, including Reinhard's function, $T_\mathrm{R}(v) =
v/(1+v)$~\citep{reinhard_high_2010}, which is used by \gls{sbmc}~\citep{gharbi_sample-based_2019},
and a variant used by Noise2Noise that adds gamma compression, $T_\mathrm{RG}(v) =
(v/(1+v))^{1/2.2}$~\citep{lehtinen_noise2noise_2018}.  Gamma compression alone, $T_\mathrm{G}(v) =
v^{1/2.2}$, is also a common tone mapping function, so we have included it in our comparisons.
These tone mapping functions, when combined with the $L_2$ or $L_\mathrm{HDR}$ losses, match the
theoretical requirements described in Section~\ref{ssec:minimizing} (again see
Figure~\ref{fig:curves}(b)), and so would be expected to perform well.  The tone mapping functions
combined with the $L_\mathrm{rMSE}$ loss have relatively high Jensen gap bounds, and so would not be
expected to perform well.

We consider three options of where to apply tone mapping: to both the model output and the target
image, to only the target image, or to neither.  Tone mapping only the target image is an
alternative version of our method that can be interpreted as asking the model to generate outputs in
the tone mapped scale.  We convert these tone mapped outputs back to the linear scale with the
inverse tone mapping function for evaluation.  We do not consider applying tone mapping to only the
model output as this would be nonsensical.  This scenario would require the model to generate huge
values that would then be tone mapped down to reach the already large \gls{hdr} values in the target
images.  \gls{sbmc} and Noise2Noise both find that tone mapping the model input $\hat{x}$ is
beneficial as well~\citep{lehtinen_noise2noise_2018,gharbi_sample-based_2019}, so we follow that
practice.  \gls{sbmc} uses a separate tone mapping function for the model input, $T_i(v) =
\frac{1}{10}\ln(1+v)$, which we keep in place for all of our experiments.

\begin{figure*}[t]
    \centering
    \includegraphics[width=\textwidth]{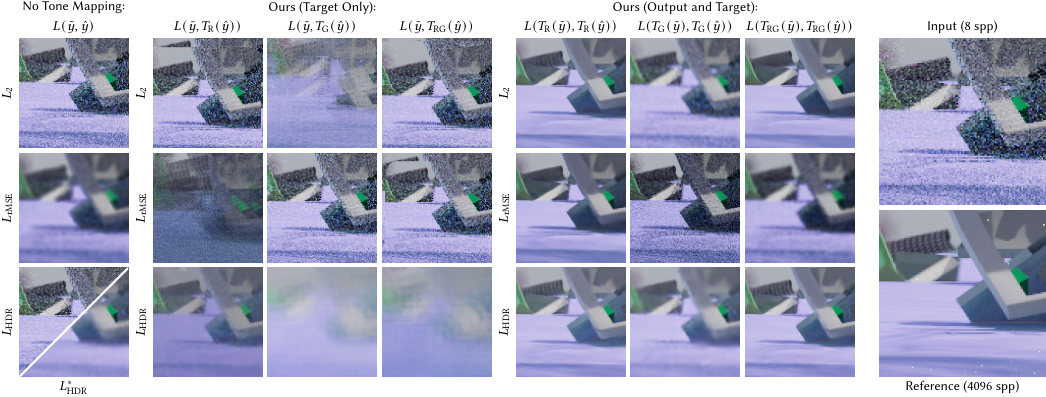}
    \caption{Model outputs for a single validation image for all combinations of loss functions,
    tone mapping functions, and tone mapping placement}
    \label{fig:tone_loss_images}
    \Description{A grid of images with 3 rows and 7 columns.  The rows are labeled L-2, L-rMSE, and
    L-HDR.  The loss functions in the first column have no tone mapping, and the bottom image in
    this column is split in half diagonally with the lower half showing the output for L-HDR Star.
    The next three columns have tone mapping on the target image only for T-R, T-G, and T-RG,
    respectively.  The last three columns have tone mapping on the model output and target images,
    again for the T-R, T-G, and T-RG tone mapping functions.  The quality of the outputs generally
    improves from left to right, with some outputs towards the right still displaying some blurring
    and noise.  To the right of the main grid are two additional images showing the noisy input and
    clean reference images.}
\end{figure*}

Evaluating all of the possible combinations of three loss functions, three tone mapping functions,
and three options of where to apply them, would require a grid search with 27 entries.  However,
some of the entries are redundant: there is no need to try the different tone mapping functions when
tone mapping is not used.  This reduces the number of trials to 21.  We also add a trial for the
$L_\mathrm{HDR}^\ast(\tilde{y},\hat{y})$ variant, bringing the total number of trials to 22.  It is
not necessary to combine $L_\mathrm{HDR}^\ast$ with tone mapping, since the bias in that case would
be the same as that of $L_2$ with tone mapping, which we already evaluate.  We trained an \gls{sbmc}
model using each of these configurations, with all other parameters remaining the same.  The
non-tone mapped \gls{rmse} loss was calculated for validation regardless of which loss function and
tone mapping function were used for training.  The training losses, gradient norms, and validation
losses for each model are shown in Figure~\ref{fig:tone_loss}.  The output of each model on a single
validation image is shown in Figure~\ref{fig:tone_loss_images}, with additional validation images
shown in Section~\ref{ssec:qualitative}.

As predicted, the plain $L_2$, $L_\mathrm{rMSE}$, and $L_\mathrm{HDR}$ loss functions do not perform
well.  $L_\mathrm{HDR}^\ast$ also performs poorly, indicating that it does not have sufficient
outlier rejection.  In Figure~\ref{fig:tone_loss}(a), these loss functions all show large and noisy
training loss values.  Note that since the training losses have different units, we cannot compare
the loss values directly.  Instead, we are interested in the numerical stability of the loss
landscapes.  In Figure~\ref{fig:tone_loss}(b), we can see that these unstable loss functions lead to
exploding and vanishing gradients in the case of $L_2(\tilde{y},\hat{y})$ and
$L_\mathrm{HDR}(\tilde{y},\hat{y})$, causing the training process to stall.  This stalled training
can be seen in Figure~\ref{fig:tone_loss}(c), where the validation losses fail to converge.
$L_\mathrm{rMSE}(\tilde{y},\hat{y})$ and $L_\mathrm{HDR}^\ast(\tilde{y},\hat{y})$ show better
stability in Figure~\ref{fig:tone_loss}(b), but have large validation loss values in
Figure~\ref{fig:tone_loss}(c).  The plain loss functions also perform poorly in
Figure~\ref{fig:tone_loss_images}, where the validation images are either overly noisy or blurry.

Turning to our method, the loss functions with tone mapping of the model output and target image
mostly perform well, as predicted.  $L_2(T(\tilde{y}),T(\hat{y}))$ and
$L_\mathrm{HDR}(T(\tilde{y}),T(\hat{y}))$ have low and smooth training losses and gradient norms in
Figures~\ref{fig:tone_loss}(a) and~\ref{fig:tone_loss}(b) for all of the tone mapping functions.
These losses also show good denoising performance in Figures~\ref{fig:tone_loss}(c)
and~\ref{fig:tone_loss_images}, with the exception of the $T_\mathrm{G}$ tone mapping function.
$T_\mathrm{R}$ and $T_\mathrm{RG}$ both perform well here, which indicates that their differing
$J(y)$ values for low values of $y$ (see Figure~\ref{fig:curves}(b)) are not important.  This
observation supports the idea that the variance bound (see Figure~\ref{fig:curves}(c)) is dominant
for low $y$.  The poor performance of $T_\mathrm{G}$ here, despite having similar $J(y)$ curves to
the other tone mapping functions, could be because of the difference shown in
Figure~\ref{fig:curves}(a), where $T_\mathrm{G}$ has unbounded output values.  The
$L_\mathrm{rMSE}(T(\tilde{y}),T(\hat{y}))$ losses perform poorly, as predicted by their large $J(y)$
values.  The loss functions with tone mapping applied to only the target image also perform poorly,
with large and noisy training losses and gradients, and poor denoising results.  The lowest
validation loss overall is achieved by
$L_\mathrm{HDR}(T_\mathrm{RG}(\tilde{y}),T_\mathrm{RG}(\hat{y}))$.  This loss function also has the
best qualitative denoising result, with more detail preserved in the background objects.

\subsection{Comparing to Clean References}
\label{ssec:comparison}

We now compare our nonlinear Noise2Noise method to training with clean reference images.  Training
with clean references is represented by the original \gls{sbmc}
implementation~\citep{gharbi_sample-based_2019}, and our method is represented by the best
performing combination of loss function and tone mapping function from
Section~\ref{ssec:model_selection}.  We did not include the models with plain loss functions or tone
mapping of only the target image in this comparison, because none of these models produced
acceptable denoising results.  We trained our model again for seven days to allow for any
performance improvements that could occur over longer training times.  We also trained another model
with the same configuration for seven days on a dataset of twice the size composed of our original
training set combined with our additional training set.  The two datasets were shuffled together so
that the model was exposed to examples from both datasets at random.  Extending the training time to
seven days resulted in slightly lower validation loss values for both models.  Training with the
larger dataset did not result in a significant improvement in validation loss compared with training
on the original dataset.

\begin{table*}[t]
    \centering  %
    \caption[]{Error metrics for each denoiser averaged over the \acrshort{sbmc} test set
    (\colorbox{red!40}{\textbf{bold}} indicates best result, \colorbox{orange!40}{\textit{italics}}
    indicates second best result)}
    \resizebox{\textwidth}{!}{%
    \begin{tabular}{llcccccccccc}\hline
                 &       & Input   & Ours~(Orig.)    & Ours~(Large)    & SBMC          & Bako            & Bako~(f.t.)     & Bitterli & Kalantari        & Rousselle        & Sen    \\ \hline
         4~spp   & rMSE  & 17.3054 & 0.8619          & 0.6130          & \best{0.0482} & 1.0867          & \second{0.4932} & 1.0847   & 1.5814           & 2.0416           & 1.0352 \\
                 & DSSIM & 0.4648  & \second{0.0815} & 0.0819          & \best{0.0685} & 0.1294          & 0.1173          & 0.1014   & 0.0869           & 0.1575           & 0.1009 \\ \hline
         8~spp   & rMSE  & 7.4732  & 0.3193          & \second{0.2940} & \best{0.0382} & 8.1238          & 0.8471          & 0.9149   & 1.6684           & 2.0721           & 0.5670 \\
                 & DSSIM & 0.3995  & 0.0707          & \second{0.0700} & \best{0.0599} & 0.1190          & 0.0940          & 0.0818   & 0.0708           & 0.1175           & 0.0987 \\ \hline
         16~spp  & rMSE  & 11.0416 & 0.1393          & \second{0.1052} & \best{0.0315} & 21.3297         & 0.2934          & 0.9488   & 1.8151           & 2.0481           & 0.3348 \\
                 & DSSIM & 0.3373  & 0.0623          & 0.0606          & \best{0.0542} & 0.1128          & 0.0762          & 0.0700   & \second{0.0600}  & 0.0898           & 0.0986 \\ \hline
         32~spp  & rMSE  & 16.5478 & 0.0841          & \second{0.0588} & \best{0.0274} & 31.4400         & 0.1427          & 1.1344   & 1.7398           & 1.6447           & 0.2731 \\
                 & DSSIM & 0.2775  & 0.0572          & 0.0545          & \best{0.0510} & 0.1106          & 0.0619          & 0.0630   & \second{0.0516}  & 0.0685           & 0.1005 \\ \hline
         64~spp  & rMSE  & 12.0608 & 0.0601          & \second{0.0437} & \best{0.0261} & 0.1359          & 0.1553          & 0.8747   & 1.6228           & 1.6974           & --     \\
                 & DSSIM & 0.2223  & 0.0547          & 0.0514          & 0.0494        & \second{0.0407} & 0.0536          & 0.0566   & \best{0.0393}    & 0.0547           & --     \\ \hline
         128~spp & rMSE  & 1.7717  & 0.0497          & \second{0.0377} & \best{0.0254} & 0.0757          & 0.1229          & 0.9039   & 1.7394           & 1.8176           & --     \\
                 & DSSIM & 0.0488  & 0.0539          & 0.0505          & 0.0488        & \best{0.0353}   & 0.0432          & 0.0565   & \second{0.0414}  & 0.0466           & --     \\ \hline
    \end{tabular}}
    \label{table:results}
\end{table*}

\setlength{\dblfloatsep}{8pt plus 4pt minus 2pt}
\begin{figure*}[t]
    \centering
    \includegraphics[width=\textwidth]{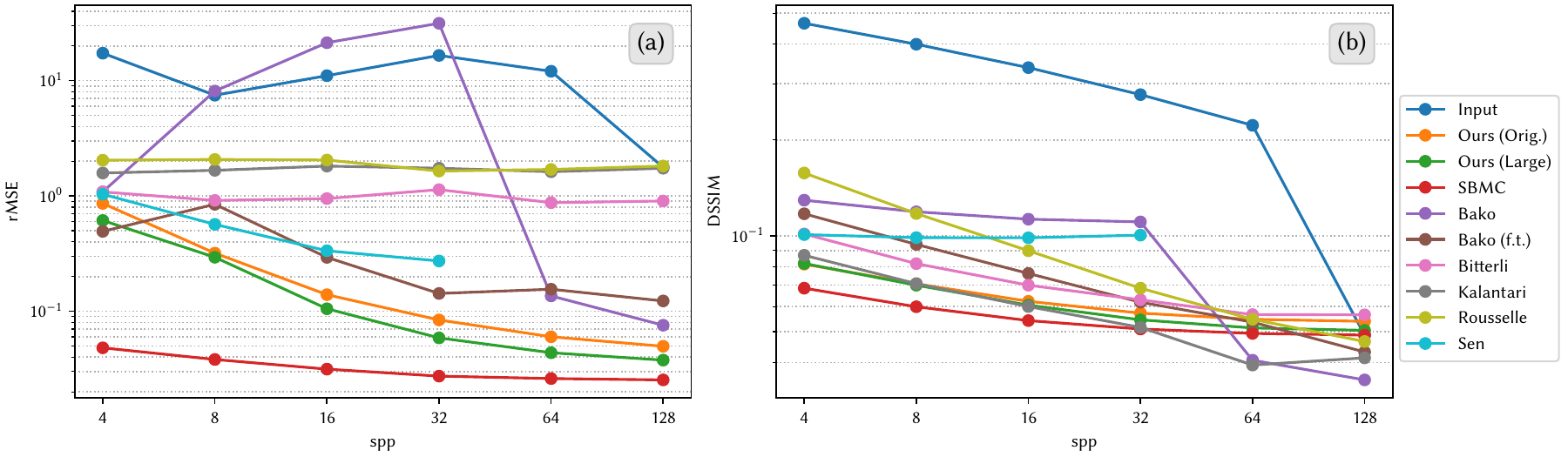}
    \caption{(a)~\acrshort{rmse} and (b)~\acrshort{dssim} error metrics for each denoiser at various
    input \acrfull{spp}, averaged over the \acrshort{sbmc} test set}
    \label{fig:results_plot}
    \Description{Two plots of error curves for the denoising models under consideration. The left
    plot has relative mean squared error on the Y axis and samples per pixel on the X axis. In this
    plot, SBMC has the lowest error values at all samples per pixel. Our models have similar error
    values at higher samples per pixel, and about an order of magnitude higher error values at lower
    samples per pixel. The other denoisers mostly have higher error values than our models.  The
    right plot has structural dissimilarity index on the Y axis and samples per pixel on the X axis.
    In this plot, SBMC has the lowest error at low samples per pixel, but is surpassed by several
    other denoisers at higher samples per pixel. Our models track SBMC closely with slightly higher
    error values at all samples per pixel.}
\end{figure*}

We performed our comparisons using the \gls{sbmc} test set, which consists of 55 scenes collected
from publicly available sources~\citep{gharbi_sample-based_2019}.  The \gls{sbmc} authors provide
reference images for each scene rendered at $1024\times1024$ resolution with
8192~\gls{spp}~\citep{gharbi_sample-based_2019}.  The low-sample count input images with additional
feature data are not provided, so we rendered these using the provided scene files.  The \gls{sbmc}
authors compare their work against five other
denoisers~\citep{bako_kernel-predicting_2017,bitterli_nonlinearly_2016,kalantari_machine_2015,rousselle_adaptive_2012,sen_filtering_2012}.
We include error numbers for \gls{sbmc} and the other denoisers using the data from Table~1 of the
\gls{sbmc} paper~\citep{gharbi_sample-based_2019}.  Comparing our models to the other denoisers in
this way is fair, since our only changes to the \gls{sbmc} training set are the lack of indoor
scenes, and rendering the references at 8~\gls{spp} instead of 4096~\gls{spp} (see
Section~\ref{ssec:methods}). These changes would only serve to disadvantage our models relative to
\gls{sbmc} and the other denoisers.  Two error metrics are used for comparison: \gls{rmse}, and
\acrlong{dssim} $\text{DSSIM} = 1 - \text{SSIM}$~\citep{wang_image_2004}.  Both of these metrics are
used without tone mapping, and are computed so that lower values are better.  The results are shown
in Table~\ref{table:results} and Figure~\ref{fig:results_plot}.  Some example outputs from our
models and \gls{sbmc} are shown in Figure~\ref{fig:comparison_full} and Section~\ref{ssec:qualitative}.  Outputs from the other denoisers are in the
\gls{sbmc} supplemental data~\citep{gharbi_sample-based_2019}.

Table~\ref{table:results} and Figure~\ref{fig:results_plot} show that the original \gls{sbmc}
implementation achieved the best \gls{rmse} results for each of the input sample counts.  \gls{sbmc}
also achieved the best \gls{dssim} results for the lower sample counts, with some of the other
denoisers taking the lead at higher sample counts.  Both of our models produced results that are
close to \gls{sbmc} for all sample counts in terms of \gls{dssim}.  The results are close for
\gls{rmse} at higher sample counts, and about an order of magnitude larger at lower sample counts.
Our model trained on the larger dataset performed better than our original model in all cases but
one.  This difference from the validation results can be explained by the validation and test sets
coming from different distributions.  Training with even more noisy data is a promising direction
for future work.  In fact, the Noise2Noise authors were able to surpass the performance of training
with clean references for a different denoising task~\citep{lehtinen_noise2noise_2018}.  Compared to
the other denoising methods, our models came in second after \gls{sbmc} in the majority of cases
where \gls{sbmc} had the best performance.  This performance was achieved despite our models being
trained only on outdoor scenes (see Section~\ref{ssec:methods}) while being evaluated on a test set
that includes many indoor scenes.  Figure~\ref{fig:comparison_full} shows outputs from our models
and \gls{sbmc} on a single test scene.  Our models compare well to \gls{sbmc}, with all three models
reducing the \gls{rmse} of the noisy input to a single-digit percentage of its original value.  In
this case, our model trained on the original dataset slightly outperforms our model trained on the
larger dataset.

\section{Limitations and Conclusions}

In this work, we showed that nonlinear functions can be effectively applied to the noisy target
images in Noise2Noise training.  We presented a theoretical framework for deriving bounds on the
bias introduced by this method, and demonstrated that the bias can be negligible compared to the
practical benefits.  To summarize: choose nonlinear functions with low curvature near the clean
target locations.  For bounded distributions, some curvature is allowable near the minimum and
maximum values.  The main limitation of our theoretical approach is that characterizing the Jensen
gap bound analytically is only possible for simple loss functions and nonlinear functions.  However,
if the bound is not tractable analytically, it may still be possible to evaluate numerically or
experimentally.  We look forward to future works that demonstrate more applications of our nonlinear
Noise2Noise method.

We demonstrated our method on the denoising of Monte Carlo rendered images.  We applied our
theoretical framework to show that certain combinations of loss functions and tone mapping functions
can reduce the effect of outliers while introducing minimal bias to the results.  Our nonlinear
Noise2Noise method makes it possible to train complex Monte Carlo denoising models for varied scenes
from only noisy training data.  Our denoising results approach those of the original implementation
trained with clean reference images, but our models are trained on $512\times$ fewer reference
samples.  We hope that future works will further improve the performance of our nonlinear method by
training with even more noisy data, possibly even exceeding the performance of training with clean
references.

\begin{acks}
Thank you to Morgan McGuire, Yuri Boykov, and the anonymous reviewers for their helpful comments.

This research was funded in part by NSERC (the Natural Sciences and Engineering Research Council of
Canada), the University of Waterloo, and David R. Cheriton.  Computing resources for this research
were provided in part by Compute Ontario and the Digital Research Alliance of Canada.
\end{acks}

\bibliographystyle{ACM-Reference-Format}
\bibliography{paper}


\begin{thebibliography}{51}


\ifx \showCODEN    \undefined \def \showCODEN     #1{\unskip}     \fi
\ifx \showISBNx    \undefined \def \showISBNx     #1{\unskip}     \fi
\ifx \showISBNxiii \undefined \def \showISBNxiii  #1{\unskip}     \fi
\ifx \showISSN     \undefined \def \showISSN      #1{\unskip}     \fi
\ifx \showLCCN     \undefined \def \showLCCN      #1{\unskip}     \fi
\ifx \shownote     \undefined \def \shownote      #1{#1}          \fi
\ifx \showarticletitle \undefined \def \showarticletitle #1{#1}   \fi
\ifx \showURL      \undefined \def \showURL       {\relax}        \fi
\providecommand\bibfield[2]{#2}
\providecommand\bibinfo[2]{#2}
\providecommand\natexlab[1]{#1}
\providecommand\showeprint[2][]{arXiv:#2}

\bibitem[Back et~al\mbox{.}(2020)]%
        {back_deep_2020}
\bibfield{author}{\bibinfo{person}{Jonghee Back}, \bibinfo{person}{Binh-Son
  Hua}, \bibinfo{person}{Toshiya Hachisuka}, {and} \bibinfo{person}{Bochang
  Moon}.} \bibinfo{year}{2020}\natexlab{}.
\newblock \showarticletitle{Deep combiner for independent and correlated pixel
  estimates}.
\newblock \bibinfo{journal}{\emph{ACM Transactions on Graphics}}
  \bibinfo{volume}{39}, \bibinfo{number}{6} (\bibinfo{date}{Nov.}
  \bibinfo{year}{2020}), \bibinfo{pages}{242:1--242:12}.
\newblock
\showISSN{0730-0301}
\href{https://doi.org/10.1145/3414685.3417847}{doi:\nolinkurl{10.1145/3414685.3417847}}


\bibitem[Back et~al\mbox{.}(2022)]%
        {back_self-supervised_2022}
\bibfield{author}{\bibinfo{person}{Jonghee Back}, \bibinfo{person}{Binh-Son
  Hua}, \bibinfo{person}{Toshiya Hachisuka}, {and} \bibinfo{person}{Bochang
  Moon}.} \bibinfo{year}{2022}\natexlab{}.
\newblock \showarticletitle{Self-{Supervised} {Post}-{Correction} for {Monte}
  {Carlo} {Denoising}}. In \bibinfo{booktitle}{\emph{{ACM} {SIGGRAPH} 2022
  {Conference} {Proceedings}}} \emph{(\bibinfo{series}{{SIGGRAPH} '22})}.
  \bibinfo{publisher}{Association for Computing Machinery},
  \bibinfo{address}{New York, NY, USA}, \bibinfo{pages}{1--8}.
\newblock
\showISBNx{978-1-4503-9337-9}
\href{https://doi.org/10.1145/3528233.3530730}{doi:\nolinkurl{10.1145/3528233.3530730}}


\bibitem[Bako et~al\mbox{.}(2017)]%
        {bako_kernel-predicting_2017}
\bibfield{author}{\bibinfo{person}{Steve Bako}, \bibinfo{person}{Thijs Vogels},
  \bibinfo{person}{Brian McWilliams}, \bibinfo{person}{Mark Meyer},
  \bibinfo{person}{Jan Novák}, \bibinfo{person}{Alex Harvill},
  \bibinfo{person}{Pradeep Sen}, \bibinfo{person}{Tony DeRose}, {and}
  \bibinfo{person}{Fabrice Rousselle}.} \bibinfo{year}{2017}\natexlab{}.
\newblock \showarticletitle{Kernel-{Predicting} {Convolutional} {Networks} for
  {Denoising} {Monte} {Carlo} {Renderings}}.
\newblock \bibinfo{journal}{\emph{ACM Transactions on Graphics (Proceedings of
  SIGGRAPH 2017)}} \bibinfo{volume}{36}, \bibinfo{number}{4}
  (\bibinfo{date}{July} \bibinfo{year}{2017}).
\newblock
\href{https://doi.org/10.1145/3072959.3073708}{doi:\nolinkurl{10.1145/3072959.3073708}}


\bibitem[Batson and Royer(2019)]%
        {batson_noise2self_2019}
\bibfield{author}{\bibinfo{person}{Joshua~D. Batson} {and}
  \bibinfo{person}{Loic~A. Royer}.} \bibinfo{year}{2019}\natexlab{}.
\newblock \showarticletitle{{Noise2Self}: {Blind} {Denoising} by
  {Self}-{Supervision}}.
\newblock \bibinfo{journal}{\emph{ICML}} (\bibinfo{year}{2019}).
\newblock
\urldef\tempurl%
\url{http://proceedings.mlr.press/v97/batson19a.html}
\showURL{%
\tempurl}


\bibitem[Bhatia and Davis(2000)]%
        {bhatia_better_2000}
\bibfield{author}{\bibinfo{person}{Rajendra Bhatia} {and}
  \bibinfo{person}{Chandler Davis}.} \bibinfo{year}{2000}\natexlab{}.
\newblock \showarticletitle{A {Better} {Bound} on the {Variance}}.
\newblock \bibinfo{journal}{\emph{The American Mathematical Monthly}}
  \bibinfo{volume}{107}, \bibinfo{number}{4} (\bibinfo{date}{April}
  \bibinfo{year}{2000}), \bibinfo{pages}{353--357}.
\newblock
\showISSN{0002-9890}
\href{https://doi.org/10.1080/00029890.2000.12005203}{doi:\nolinkurl{10.1080/00029890.2000.12005203}}
\newblock
\shownote{Publisher: Taylor \& Francis}.


\bibitem[Bitterli et~al\mbox{.}(2016)]%
        {bitterli_nonlinearly_2016}
\bibfield{author}{\bibinfo{person}{Benedikt Bitterli}, \bibinfo{person}{Fabrice
  Rousselle}, \bibinfo{person}{Bochang Moon}, \bibinfo{person}{José~A.
  Iglesias-Guitián}, \bibinfo{person}{David Adler}, \bibinfo{person}{Kenny
  Mitchell}, \bibinfo{person}{Wojciech Jarosz}, {and} \bibinfo{person}{Jan
  Novák}.} \bibinfo{year}{2016}\natexlab{}.
\newblock \showarticletitle{Nonlinearly {Weighted} {First}-order {Regression}
  for {Denoising} {Monte} {Carlo} {Renderings}}.
\newblock \bibinfo{journal}{\emph{Computer Graphics Forum}}
  \bibinfo{volume}{35}, \bibinfo{number}{4} (\bibinfo{date}{July}
  \bibinfo{year}{2016}), \bibinfo{pages}{107--117}.
\newblock
\showISSN{0167-7055}
\href{https://doi.org/10.1111/cgf.12954}{doi:\nolinkurl{10.1111/cgf.12954}}


\bibitem[Bora et~al\mbox{.}(2018)]%
        {bora_ambientgan_2018}
\bibfield{author}{\bibinfo{person}{Ashish Bora}, \bibinfo{person}{Eric Price},
  {and} \bibinfo{person}{{Alexandros G. Dimakis}}.}
  \bibinfo{year}{2018}\natexlab{}.
\newblock \showarticletitle{{AmbientGAN}: {Generative} models from lossy
  measurements}. In \bibinfo{booktitle}{\emph{{ICLR}}}.
\newblock
\urldef\tempurl%
\url{https://openreview.net/forum?id=Hy7fDog0b}
\showURL{%
\tempurl}


\bibitem[Buades et~al\mbox{.}(2005)]%
        {buades_non-local_2005}
\bibfield{author}{\bibinfo{person}{Antoni Buades}, \bibinfo{person}{Bartomeu
  Coll}, {and} \bibinfo{person}{Jean-Michel Morel}.}
  \bibinfo{year}{2005}\natexlab{}.
\newblock \showarticletitle{A non-local algorithm for image denoising}. In
  \bibinfo{booktitle}{\emph{2005 {IEEE} {Computer} {Society} {Conference} on
  {Computer} {Vision} and {Pattern} {Recognition} ({CVPR}'05)}},
  Vol.~\bibinfo{volume}{2}. \bibinfo{publisher}{IEEE}, \bibinfo{pages}{60--65}.
\newblock
\href{https://doi.org/10.1109/CVPR.2005.38}{doi:\nolinkurl{10.1109/CVPR.2005.38}}


\bibitem[Chaitanya et~al\mbox{.}(2017)]%
        {chaitanya_interactive_2017}
\bibfield{author}{\bibinfo{person}{Chakravarty R.~Alla Chaitanya},
  \bibinfo{person}{Anton~S. Kaplanyan}, \bibinfo{person}{Christoph Schied},
  \bibinfo{person}{Marco Salvi}, \bibinfo{person}{Aaron Lefohn},
  \bibinfo{person}{Derek Nowrouzezahrai}, {and} \bibinfo{person}{Timo Aila}.}
  \bibinfo{year}{2017}\natexlab{}.
\newblock \showarticletitle{Interactive reconstruction of {Monte} {Carlo} image
  sequences using a recurrent denoising autoencoder}.
\newblock \bibinfo{journal}{\emph{ACM Transactions on Graphics (TOG)}}
  \bibinfo{volume}{36}, \bibinfo{number}{4} (\bibinfo{year}{2017}),
  \bibinfo{pages}{98}.
\newblock
\href{https://doi.org/10.1145/3072959.3073601}{doi:\nolinkurl{10.1145/3072959.3073601}}


\bibitem[Chen et~al\mbox{.}(2023)]%
        {chen_monte_2023}
\bibfield{author}{\bibinfo{person}{Bingyi Chen}, \bibinfo{person}{Zengyu Liu},
  \bibinfo{person}{Li Yuan}, \bibinfo{person}{Zhitao Liu}, \bibinfo{person}{Yi
  Li}, \bibinfo{person}{Guan Wang}, {and} \bibinfo{person}{Ning Xie}.}
  \bibinfo{year}{2023}\natexlab{}.
\newblock \showarticletitle{Monte {Carlo} {Denoising} via {Multi}-scale
  {Auxiliary} {Feature} {Fusion} {Guided} {Transformer}}. In
  \bibinfo{booktitle}{\emph{{SIGGRAPH} {Asia} 2023 {Technical}
  {Communications}}} \emph{(\bibinfo{series}{{SA} '23})}.
  \bibinfo{publisher}{Association for Computing Machinery},
  \bibinfo{address}{New York, NY, USA}, \bibinfo{pages}{1--4}.
\newblock
\showISBNx{979-8-4007-0314-0}
\href{https://doi.org/10.1145/3610543.3626179}{doi:\nolinkurl{10.1145/3610543.3626179}}


\bibitem[Chen et~al\mbox{.}(2024)]%
        {chen_temporally_2024}
\bibfield{author}{\bibinfo{person}{Chuhao Chen}, \bibinfo{person}{Yuze He},
  {and} \bibinfo{person}{Tzu-Mao Li}.} \bibinfo{year}{2024}\natexlab{}.
\newblock \showarticletitle{Temporally {Stable} {Metropolis} {Light}
  {Transport} {Denoising} using {Recurrent} {Transformer} {Blocks}}.
\newblock \bibinfo{journal}{\emph{ACM Trans. Graph.}} \bibinfo{volume}{43},
  \bibinfo{number}{4} (\bibinfo{date}{July} \bibinfo{year}{2024}),
  \bibinfo{pages}{123:1--123:14}.
\newblock
\showISSN{0730-0301}
\href{https://doi.org/10.1145/3658218}{doi:\nolinkurl{10.1145/3658218}}


\bibitem[Cho et~al\mbox{.}(2021)]%
        {cho_weakly-supervised_2021}
\bibfield{author}{\bibinfo{person}{In-Young Cho}, \bibinfo{person}{Yuchi Huo},
  {and} \bibinfo{person}{Sung-Eui Yoon}.} \bibinfo{year}{2021}\natexlab{}.
\newblock \showarticletitle{Weakly-supervised contrastive learning in path
  manifold for {Monte} {Carlo} image reconstruction}.
\newblock \bibinfo{journal}{\emph{ACM Transactions on Graphics}}
  \bibinfo{volume}{40}, \bibinfo{number}{4} (\bibinfo{date}{July}
  \bibinfo{year}{2021}), \bibinfo{pages}{38:1--38:14}.
\newblock
\showISSN{0730-0301}
\href{https://doi.org/10.1145/3450626.3459876}{doi:\nolinkurl{10.1145/3450626.3459876}}


\bibitem[Choi et~al\mbox{.}(2024)]%
        {choi_online_2024}
\bibfield{author}{\bibinfo{person}{Hajin Choi}, \bibinfo{person}{Seokpyo Hong},
  \bibinfo{person}{Inwoo Ha}, \bibinfo{person}{Nahyup Kang}, {and}
  \bibinfo{person}{Bochang Moon}.} \bibinfo{year}{2024}\natexlab{}.
\newblock \showarticletitle{Online {Neural} {Denoising} with
  {Cross}-{Regression} for {Interactive} {Rendering}}.
\newblock \bibinfo{journal}{\emph{ACM Trans. Graph.}} \bibinfo{volume}{43},
  \bibinfo{number}{6} (\bibinfo{date}{Nov.} \bibinfo{year}{2024}),
  \bibinfo{pages}{221:1--221:12}.
\newblock
\showISSN{0730-0301}
\href{https://doi.org/10.1145/3687938}{doi:\nolinkurl{10.1145/3687938}}


\bibitem[Dabov et~al\mbox{.}(2007)]%
        {dabov_image_2007}
\bibfield{author}{\bibinfo{person}{Kostadin Dabov}, \bibinfo{person}{Alessandro
  Foi}, \bibinfo{person}{Vladimir Katkovnik}, {and} \bibinfo{person}{Karen
  Egiazarian}.} \bibinfo{year}{2007}\natexlab{}.
\newblock \showarticletitle{Image {Denoising} by {Sparse} 3-{D}
  {Transform}-{Domain} {Collaborative} {Filtering}}.
\newblock \bibinfo{journal}{\emph{IEEE Transactions on Image Processing}}
  \bibinfo{volume}{16}, \bibinfo{number}{8} (\bibinfo{date}{Aug.}
  \bibinfo{year}{2007}), \bibinfo{pages}{2080--2095}.
\newblock
\showISSN{1941-0042}
\href{https://doi.org/10.1109/TIP.2007.901238}{doi:\nolinkurl{10.1109/TIP.2007.901238}}


\bibitem[Gharbi et~al\mbox{.}(2019)]%
        {gharbi_sample-based_2019}
\bibfield{author}{\bibinfo{person}{Michaël Gharbi}, \bibinfo{person}{Tzu-Mao
  Li}, \bibinfo{person}{Miika Aittala}, \bibinfo{person}{Jaakko Lehtinen},
  {and} \bibinfo{person}{Frédo Durand}.} \bibinfo{year}{2019}\natexlab{}.
\newblock \showarticletitle{Sample-based {Monte} {Carlo} {Denoising} {Using} a
  {Kernel}-splatting {Network}}.
\newblock \bibinfo{journal}{\emph{ACM Trans. Graph.}} \bibinfo{volume}{38},
  \bibinfo{number}{4} (\bibinfo{date}{July} \bibinfo{year}{2019}),
  \bibinfo{pages}{125:1--125:12}.
\newblock
\showISSN{0730-0301}
\href{https://doi.org/10.1145/3306346.3322954}{doi:\nolinkurl{10.1145/3306346.3322954}}


\bibitem[Guo et~al\mbox{.}(2019)]%
        {guo_gradnet_2019}
\bibfield{author}{\bibinfo{person}{Jie Guo}, \bibinfo{person}{Mengtian Li},
  \bibinfo{person}{Quewei Li}, \bibinfo{person}{Yuting Qiang},
  \bibinfo{person}{Bingyang Hu}, \bibinfo{person}{Yanwen Guo}, {and}
  \bibinfo{person}{Ling-Qi Yan}.} \bibinfo{year}{2019}\natexlab{}.
\newblock \showarticletitle{{GradNet}: unsupervised deep screened poisson
  reconstruction for gradient-domain rendering}.
\newblock \bibinfo{journal}{\emph{ACM Transactions on Graphics}}
  \bibinfo{volume}{38}, \bibinfo{number}{6} (\bibinfo{date}{Nov.}
  \bibinfo{year}{2019}), \bibinfo{pages}{223:1--223:13}.
\newblock
\showISSN{0730-0301}
\href{https://doi.org/10.1145/3355089.3356538}{doi:\nolinkurl{10.1145/3355089.3356538}}


\bibitem[Huo and Yoon(2021)]%
        {huo_survey_2021}
\bibfield{author}{\bibinfo{person}{Yuchi Huo} {and} \bibinfo{person}{Sung-eui
  Yoon}.} \bibinfo{year}{2021}\natexlab{}.
\newblock \showarticletitle{A survey on deep learning-based {Monte} {Carlo}
  denoising}.
\newblock \bibinfo{journal}{\emph{Computational Visual Media}}
  \bibinfo{volume}{7}, \bibinfo{number}{2} (\bibinfo{date}{June}
  \bibinfo{year}{2021}), \bibinfo{pages}{169--185}.
\newblock
\showISSN{2096-0662}
\href{https://doi.org/10.1007/s41095-021-0209-9}{doi:\nolinkurl{10.1007/s41095-021-0209-9}}


\bibitem[Jensen(1906)]%
        {jensen_sur_1906}
\bibfield{author}{\bibinfo{person}{J.~L. W.~V. Jensen}.}
  \bibinfo{year}{1906}\natexlab{}.
\newblock \showarticletitle{Sur les fonctions convexes et les inégalités
  entre les valeurs moyennes}.
\newblock \bibinfo{journal}{\emph{Acta Mathematica}}  \bibinfo{volume}{30}
  (\bibinfo{date}{Jan.} \bibinfo{year}{1906}), \bibinfo{pages}{175--193}.
\newblock
\showISSN{0001-5962, 1871-2509}
\href{https://doi.org/10.1007/BF02418571}{doi:\nolinkurl{10.1007/BF02418571}}
\newblock
\shownote{Publisher: Institut Mittag-Leffler}.


\bibitem[Kalantari et~al\mbox{.}(2015)]%
        {kalantari_machine_2015}
\bibfield{author}{\bibinfo{person}{Nima~Khademi Kalantari},
  \bibinfo{person}{Steve Bako}, {and} \bibinfo{person}{Pradeep Sen}.}
  \bibinfo{year}{2015}\natexlab{}.
\newblock \showarticletitle{A {Machine} {Learning} {Approach} for {Filtering}
  {Monte} {Carlo} {Noise}}.
\newblock \bibinfo{journal}{\emph{ACM Trans. Graph.}} \bibinfo{volume}{34},
  \bibinfo{number}{4} (\bibinfo{date}{July} \bibinfo{year}{2015}),
  \bibinfo{pages}{122:1--122:12}.
\newblock
\showISSN{0730-0301}
\href{https://doi.org/10.1145/2766977}{doi:\nolinkurl{10.1145/2766977}}


\bibitem[Krull et~al\mbox{.}(2019)]%
        {krull_noise2void_2019}
\bibfield{author}{\bibinfo{person}{Alexander Krull},
  \bibinfo{person}{Tim-Oliver Buchholz}, {and} \bibinfo{person}{Florian Jug}.}
  \bibinfo{year}{2019}\natexlab{}.
\newblock \showarticletitle{{Noise2Void} - {Learning} {Denoising} {From}
  {Single} {Noisy} {Images}}. In \bibinfo{booktitle}{\emph{Proceedings of the
  {IEEE}/{CVF} {Conference} on {Computer} {Vision} and {Pattern} {Recognition}
  ({CVPR})}}. \bibinfo{pages}{2129--2137}.
\newblock
\href{https://doi.org/10.1109/CVPR.2019.00223}{doi:\nolinkurl{10.1109/CVPR.2019.00223}}


\bibitem[Laine et~al\mbox{.}(2019)]%
        {laine_high-quality_2019}
\bibfield{author}{\bibinfo{person}{Samuli Laine}, \bibinfo{person}{Tero
  Karras}, \bibinfo{person}{Jaakko Lehtinen}, {and} \bibinfo{person}{Timo
  Aila}.} \bibinfo{year}{2019}\natexlab{}.
\newblock \showarticletitle{High-{Quality} {Self}-{Supervised} {Deep} {Image}
  {Denoising}}. In \bibinfo{booktitle}{\emph{Advances in {Neural} {Information}
  {Processing} {Systems} 32 ({NeurIPS} 2019)}}, Vol.~\bibinfo{volume}{32}.
  \bibinfo{pages}{6970--6980}.
\newblock
\urldef\tempurl%
\url{https://proceedings.neurips.cc/paper/2019/hash/2119b8d43eafcf353e07d7cb5554170b-Abstract.html}
\showURL{%
\tempurl}


\bibitem[Lee et~al\mbox{.}(2021)]%
        {lee_further_2021}
\bibfield{author}{\bibinfo{person}{Sang~Kyu Lee}, \bibinfo{person}{Jae~Ho
  Chang}, {and} \bibinfo{person}{Hyoung-Moon Kim}.}
  \bibinfo{year}{2021}\natexlab{}.
\newblock \showarticletitle{Further sharpening of {Jensen}'s inequality}.
\newblock \bibinfo{journal}{\emph{Statistics}} \bibinfo{volume}{55},
  \bibinfo{number}{5} (\bibinfo{date}{Sept.} \bibinfo{year}{2021}),
  \bibinfo{pages}{1154--1168}.
\newblock
\showISSN{0233-1888}
\href{https://doi.org/10.1080/02331888.2021.1998052}{doi:\nolinkurl{10.1080/02331888.2021.1998052}}


\bibitem[Lehtinen et~al\mbox{.}(2018)]%
        {lehtinen_noise2noise_2018}
\bibfield{author}{\bibinfo{person}{Jaakko Lehtinen}, \bibinfo{person}{Jacob
  Munkberg}, \bibinfo{person}{Jon Hasselgren}, \bibinfo{person}{Samuli Laine},
  \bibinfo{person}{Tero Karras}, \bibinfo{person}{Miika Aittala}, {and}
  \bibinfo{person}{Timo Aila}.} \bibinfo{year}{2018}\natexlab{}.
\newblock \showarticletitle{{Noise2Noise}: {Learning} {Image} {Restoration}
  without {Clean} {Data}}. In \bibinfo{booktitle}{\emph{International
  {Conference} on {Machine} {Learning}}}. \bibinfo{publisher}{PMLR},
  \bibinfo{pages}{2965--2974}.
\newblock
\urldef\tempurl%
\url{http://proceedings.mlr.press/v80/lehtinen18a.html}
\showURL{%
\tempurl}


\bibitem[Liao and Berg(2019)]%
        {liao_sharpening_2019}
\bibfield{author}{\bibinfo{person}{J.~G. Liao} {and} \bibinfo{person}{Arthur
  Berg}.} \bibinfo{year}{2019}\natexlab{}.
\newblock \showarticletitle{Sharpening {Jensen}'s {Inequality}}.
\newblock \bibinfo{journal}{\emph{The American Statistician}}
  \bibinfo{volume}{73}, \bibinfo{number}{3} (\bibinfo{date}{July}
  \bibinfo{year}{2019}), \bibinfo{pages}{278--281}.
\newblock
\showISSN{0003-1305}
\href{https://doi.org/10.1080/00031305.2017.1419145}{doi:\nolinkurl{10.1080/00031305.2017.1419145}}
\newblock
\shownote{Publisher: American Statistical Association}.


\bibitem[Lin et~al\mbox{.}(2020)]%
        {lin_detail_2020}
\bibfield{author}{\bibinfo{person}{Weiheng Lin}, \bibinfo{person}{Beibei Wang},
  \bibinfo{person}{Lu Wang}, {and} \bibinfo{person}{Nicolas Holzschuch}.}
  \bibinfo{year}{2020}\natexlab{}.
\newblock \showarticletitle{A {Detail} {Preserving} {Neural} {Network} {Model}
  for {Monte} {Carlo} {Denoising}}.
\newblock \bibinfo{journal}{\emph{Computational Visual Media}}
  \bibinfo{volume}{6} (\bibinfo{date}{April} \bibinfo{year}{2020}),
  \bibinfo{pages}{157--168}.
\newblock
\href{https://doi.org/10.1007/s41095-020-0167-7}{doi:\nolinkurl{10.1007/s41095-020-0167-7}}


\bibitem[Lin et~al\mbox{.}(2021)]%
        {lin_path-based_2021}
\bibfield{author}{\bibinfo{person}{Weiheng Lin}, \bibinfo{person}{Beibei Wang},
  \bibinfo{person}{Jian Yang}, \bibinfo{person}{Lu Wang}, {and}
  \bibinfo{person}{Ling-Qi Yan}.} \bibinfo{year}{2021}\natexlab{}.
\newblock \showarticletitle{Path-based {Monte} {Carlo} {Denoising} {Using} a
  {Three}-{Scale} {Neural} {Network}}.
\newblock \bibinfo{journal}{\emph{Computer Graphics Forum}}
  \bibinfo{volume}{40}, \bibinfo{number}{1} (\bibinfo{date}{Feb.}
  \bibinfo{year}{2021}), \bibinfo{pages}{369--381}.
\newblock
\showISSN{1467-8659}
\href{https://doi.org/10.1111/cgf.14194}{doi:\nolinkurl{10.1111/cgf.14194}}


\bibitem[Lu et~al\mbox{.}(2021)]%
        {lu_denoising_2021}
\bibfield{author}{\bibinfo{person}{Yifan Lu}, \bibinfo{person}{Siyuan Fu},
  \bibinfo{person}{Xiao~Hua Zhang}, {and} \bibinfo{person}{Ning Xie}.}
  \bibinfo{year}{2021}\natexlab{}.
\newblock \showarticletitle{Denoising {Monte} {Carlo} renderings via a
  multi-scale featured dual-residual {GAN}}.
\newblock \bibinfo{journal}{\emph{The Visual Computer}} \bibinfo{volume}{37},
  \bibinfo{number}{9} (\bibinfo{date}{Sept.} \bibinfo{year}{2021}),
  \bibinfo{pages}{2513--2525}.
\newblock
\showISSN{1432-2315}
\href{https://doi.org/10.1007/s00371-021-02204-4}{doi:\nolinkurl{10.1007/s00371-021-02204-4}}


\bibitem[Lu et~al\mbox{.}(2020)]%
        {lu_dmcr-gan_2020}
\bibfield{author}{\bibinfo{person}{Yifan Lu}, \bibinfo{person}{Ning Xie}, {and}
  \bibinfo{person}{Heng~Tao Shen}.} \bibinfo{year}{2020}\natexlab{}.
\newblock \showarticletitle{{DMCR}-{GAN}: {Adversarial} {Denoising} for {Monte}
  {Carlo} {Renderings} with {Residual} {Attention} {Networks} and
  {Hierarchical} {Features} {Modulation} of {Auxiliary} {Buffers}}. In
  \bibinfo{booktitle}{\emph{{SIGGRAPH} {Asia} 2020 {Technical}
  {Communications}}}. \bibinfo{publisher}{Association for Computing Machinery},
  \bibinfo{pages}{1--4}.
\newblock
\showISBNx{978-1-4503-8080-5}
\href{https://doi.org/10.1145/3410700.3425426}{doi:\nolinkurl{10.1145/3410700.3425426}}


\bibitem[Mangasarian(1965)]%
        {mangasarian_pseudo-convex_1965}
\bibfield{author}{\bibinfo{person}{O.~L. Mangasarian}.}
  \bibinfo{year}{1965}\natexlab{}.
\newblock \showarticletitle{Pseudo-{Convex} {Functions}}.
\newblock \bibinfo{journal}{\emph{Journal of the Society for Industrial and
  Applied Mathematics Series A Control}} \bibinfo{volume}{3},
  \bibinfo{number}{2} (\bibinfo{date}{Jan.} \bibinfo{year}{1965}),
  \bibinfo{pages}{281--290}.
\newblock
\showISSN{0887-4603}
\href{https://doi.org/10.1137/0303020}{doi:\nolinkurl{10.1137/0303020}}
\newblock
\shownote{Publisher: Society for Industrial and Applied Mathematics}.


\bibitem[Meng et~al\mbox{.}(2020)]%
        {meng_real-time_2020}
\bibfield{author}{\bibinfo{person}{Xiaoxu Meng}, \bibinfo{person}{Quan Zheng},
  \bibinfo{person}{Amitabh Varshney}, \bibinfo{person}{Gurprit Singh}, {and}
  \bibinfo{person}{Matthias Zwicker}.} \bibinfo{year}{2020}\natexlab{}.
\newblock \showarticletitle{Real-time {Monte} {Carlo} {Denoising} with the
  {Neural} {Bilateral} {Grid}}. In \bibinfo{booktitle}{\emph{Eurographics
  {Symposium} on {Rendering} ({EGSR})}}. \bibinfo{publisher}{The Eurographics
  Association}.
\newblock
\href{https://doi.org/10.2312/sr.20201133}{doi:\nolinkurl{10.2312/sr.20201133}}


\bibitem[Munkberg and Hasselgren(2020)]%
        {munkberg_neural_2020}
\bibfield{author}{\bibinfo{person}{Jacob Munkberg} {and} \bibinfo{person}{Jon
  Hasselgren}.} \bibinfo{year}{2020}\natexlab{}.
\newblock \showarticletitle{Neural {Denoising} with {Layer} {Embeddings}}. In
  \bibinfo{booktitle}{\emph{Computer {Graphics} {Forum}}},
  Vol.~\bibinfo{volume}{39}. \bibinfo{pages}{1--12}.
\newblock
\href{https://doi.org/10.1111/cgf.14049}{doi:\nolinkurl{10.1111/cgf.14049}}


\bibitem[Oh and Moon(2024)]%
        {oh_joint_2024}
\bibfield{author}{\bibinfo{person}{Geunwoo Oh} {and} \bibinfo{person}{Bochang
  Moon}.} \bibinfo{year}{2024}\natexlab{}.
\newblock \showarticletitle{Joint self-attention for denoising {Monte} {Carlo}
  rendering}.
\newblock \bibinfo{journal}{\emph{The Visual Computer}} \bibinfo{volume}{40},
  \bibinfo{number}{7} (\bibinfo{date}{July} \bibinfo{year}{2024}),
  \bibinfo{pages}{4623--4634}.
\newblock
\showISSN{1432-2315}
\href{https://doi.org/10.1007/s00371-024-03446-8}{doi:\nolinkurl{10.1007/s00371-024-03446-8}}


\bibitem[Pharr et~al\mbox{.}(2023)]%
        {pharr_physically_2023}
\bibfield{author}{\bibinfo{person}{Matt Pharr}, \bibinfo{person}{Wenzel Jakob},
  {and} \bibinfo{person}{Greg Humphreys}.} \bibinfo{year}{2023}\natexlab{}.
\newblock \bibinfo{booktitle}{\emph{Physically {Based} {Rendering}, fourth
  edition: {From} {Theory} to {Implementation}}}.
\newblock \bibinfo{publisher}{MIT Press}.
\newblock
\showISBNx{978-0-262-04802-6}
\urldef\tempurl%
\url{https://pbrt.org/}
\showURL{%
\tempurl}


\bibitem[Popoviciu(1935)]%
        {popoviciu_sur_1935}
\bibfield{author}{\bibinfo{person}{Tiberiu Popoviciu}.}
  \bibinfo{year}{1935}\natexlab{}.
\newblock \showarticletitle{Sur les équations algébriques ayant toutes leurs
  racines réelles}.
\newblock \bibinfo{journal}{\emph{Mathematica}} \bibinfo{volume}{9},
  \bibinfo{number}{129-145} (\bibinfo{year}{1935}), \bibinfo{pages}{20}.
\newblock


\bibitem[Reinhard et~al\mbox{.}(2010)]%
        {reinhard_high_2010}
\bibfield{author}{\bibinfo{person}{Erik Reinhard}, \bibinfo{person}{Wolfgang
  Heidrich}, \bibinfo{person}{Paul Debevec}, \bibinfo{person}{Sumanta
  Pattanaik}, \bibinfo{person}{Greg Ward}, {and} \bibinfo{person}{Karol
  Myszkowski}.} \bibinfo{year}{2010}\natexlab{}.
\newblock \bibinfo{booktitle}{\emph{High {Dynamic} {Range} {Imaging}:
  {Acquisition}, {Display}, and {Image}-{Based} {Lighting}}}.
\newblock \bibinfo{publisher}{Morgan Kaufmann}.
\newblock
\showISBNx{978-0-08-095711-1}


\bibitem[Robitzski(2019)]%
        {robitzski_startup_2019}
\bibfield{author}{\bibinfo{person}{Dan Robitzski}.}
  \bibinfo{year}{2019}\natexlab{}.
\newblock \bibinfo{title}{A startup is suing {Facebook}, {Princeton} for
  stealing its {AI} data}.
\newblock
\urldef\tempurl%
\url{https://futurism.com/tech-suing-facebook-princeton-data}
\showURL{%
\tempurl}


\bibitem[Rousselle et~al\mbox{.}(2011)]%
        {rousselle_adaptive_2011}
\bibfield{author}{\bibinfo{person}{Fabrice Rousselle}, \bibinfo{person}{Claude
  Knaus}, {and} \bibinfo{person}{Matthias Zwicker}.}
  \bibinfo{year}{2011}\natexlab{}.
\newblock \showarticletitle{Adaptive sampling and reconstruction using greedy
  error minimization}. In \bibinfo{booktitle}{\emph{{ACM} {Transactions} on
  {Graphics} ({TOG})}}, Vol.~\bibinfo{volume}{30}. \bibinfo{publisher}{ACM},
  \bibinfo{pages}{159}.
\newblock
\href{https://doi.org/10.1145/2024156.2024193}{doi:\nolinkurl{10.1145/2024156.2024193}}


\bibitem[Rousselle et~al\mbox{.}(2012)]%
        {rousselle_adaptive_2012}
\bibfield{author}{\bibinfo{person}{Fabrice Rousselle}, \bibinfo{person}{Claude
  Knaus}, {and} \bibinfo{person}{Matthias Zwicker}.}
  \bibinfo{year}{2012}\natexlab{}.
\newblock \showarticletitle{Adaptive rendering with non-local means filtering}.
\newblock \bibinfo{journal}{\emph{ACM Transactions on Graphics}}
  \bibinfo{volume}{31}, \bibinfo{number}{6} (\bibinfo{date}{Nov.}
  \bibinfo{year}{2012}), \bibinfo{pages}{195:1--195:11}.
\newblock
\showISSN{0730-0301}
\href{https://doi.org/10.1145/2366145.2366214}{doi:\nolinkurl{10.1145/2366145.2366214}}


\bibitem[Sen and Darabi(2012)]%
        {sen_filtering_2012}
\bibfield{author}{\bibinfo{person}{Pradeep Sen} {and} \bibinfo{person}{Soheil
  Darabi}.} \bibinfo{year}{2012}\natexlab{}.
\newblock \showarticletitle{On filtering the noise from the random parameters
  in {Monte} {Carlo} rendering}.
\newblock \bibinfo{journal}{\emph{ACM Transactions on Graphics}}
  \bibinfo{volume}{31}, \bibinfo{number}{3} (\bibinfo{date}{May}
  \bibinfo{year}{2012}), \bibinfo{pages}{1--15}.
\newblock
\href{https://doi.org/10.1145/2167076.2167083}{doi:\nolinkurl{10.1145/2167076.2167083}}


\bibitem[Song et~al\mbox{.}(2017)]%
        {song_semantic_2017}
\bibfield{author}{\bibinfo{person}{Shuran Song}, \bibinfo{person}{Fisher Yu},
  \bibinfo{person}{Andy Zeng}, \bibinfo{person}{Angel~X. Chang},
  \bibinfo{person}{Manolis Savva}, {and} \bibinfo{person}{Thomas Funkhouser}.}
  \bibinfo{year}{2017}\natexlab{}.
\newblock \showarticletitle{Semantic {Scene} {Completion} from a {Single}
  {Depth} {Image}}. In \bibinfo{booktitle}{\emph{Proceedings of the {IEEE}
  {Conference} on {Computer} {Vision} and {Pattern} {Recognition} ({CVPR})}}.
  \bibinfo{pages}{1746--1754}.
\newblock
\href{https://doi.org/10.1109/CVPR.2017.28}{doi:\nolinkurl{10.1109/CVPR.2017.28}}


\bibitem[Tinits(2022)]%
        {tinits_learning_2022}
\bibfield{author}{\bibinfo{person}{Andrew Tinits}.}
  \bibinfo{year}{2022}\natexlab{}.
\newblock \emph{\bibinfo{title}{Learning {Sample}-{Based} {Monte} {Carlo}
  {Denoising} from {Noisy} {Training} {Data}}}.
\newblock \bibinfo{thesistype}{Master's\ thesis}. \bibinfo{school}{University
  of Waterloo}, \bibinfo{address}{Waterloo, Ontario, Canada}.
\newblock
\urldef\tempurl%
\url{http://hdl.handle.net/10012/18071}
\showURL{%
\tempurl}


\bibitem[Vavilala et~al\mbox{.}(2024)]%
        {vavilala_denoising_2024}
\bibfield{author}{\bibinfo{person}{Vaibhav Vavilala}, \bibinfo{person}{Rahul
  Vasanth}, {and} \bibinfo{person}{David Forsyth}.}
  \bibinfo{year}{2024}\natexlab{}.
\newblock \bibinfo{title}{Denoising {Monte} {Carlo} {Renders} {With}
  {Diffusion} {Models}}.
\newblock
\href{https://doi.org/10.48550/arXiv.2404.00491}{doi:\nolinkurl{10.48550/arXiv.2404.00491}}


\bibitem[Vogels et~al\mbox{.}(2018)]%
        {vogels_denoising_2018}
\bibfield{author}{\bibinfo{person}{Thijs Vogels}, \bibinfo{person}{Fabrice
  Rousselle}, \bibinfo{person}{Brian McWilliams}, \bibinfo{person}{Gerhard
  Röthlin}, \bibinfo{person}{Alex Harvill}, \bibinfo{person}{David Adler},
  \bibinfo{person}{Mark Meyer}, {and} \bibinfo{person}{Jan Novák}.}
  \bibinfo{year}{2018}\natexlab{}.
\newblock \showarticletitle{Denoising with {Kernel} {Prediction} and
  {Asymmetric} {Loss} {Functions}}.
\newblock \bibinfo{journal}{\emph{ACM Transactions on Graphics}}
  \bibinfo{volume}{37}, \bibinfo{number}{4} (\bibinfo{date}{July}
  \bibinfo{year}{2018}), \bibinfo{pages}{124:1--124:15}.
\newblock
\showISSN{0730-0301}
\href{https://doi.org/10.1145/3197517.3201388}{doi:\nolinkurl{10.1145/3197517.3201388}}


\bibitem[Wang et~al\mbox{.}(2004)]%
        {wang_image_2004}
\bibfield{author}{\bibinfo{person}{Zhou Wang}, \bibinfo{person}{Alan~Conrad
  Bovik}, \bibinfo{person}{Hamid~Rahim Sheikh}, {and} \bibinfo{person}{Eero~P.
  Simoncelli}.} \bibinfo{year}{2004}\natexlab{}.
\newblock \showarticletitle{Image quality assessment: from error visibility to
  structural similarity}.
\newblock \bibinfo{journal}{\emph{IEEE Transactions on Image Processing}}
  \bibinfo{volume}{13}, \bibinfo{number}{4} (\bibinfo{date}{April}
  \bibinfo{year}{2004}), \bibinfo{pages}{600--612}.
\newblock
\href{https://doi.org/10.1109/TIP.2003.819861}{doi:\nolinkurl{10.1109/TIP.2003.819861}}


\bibitem[Wong and Wong(2019)]%
        {wong_deep_2019}
\bibfield{author}{\bibinfo{person}{Kin-Ming Wong} {and}
  \bibinfo{person}{Tien-Tsin Wong}.} \bibinfo{year}{2019}\natexlab{}.
\newblock \showarticletitle{Deep residual learning for denoising {Monte}
  {Carlo} renderings}.
\newblock \bibinfo{journal}{\emph{Computational Visual Media}}
  \bibinfo{volume}{5}, \bibinfo{number}{3} (\bibinfo{date}{Sept.}
  \bibinfo{year}{2019}), \bibinfo{pages}{239--255}.
\newblock
\showISSN{2096-0662}
\href{https://doi.org/10.1007/s41095-019-0142-3}{doi:\nolinkurl{10.1007/s41095-019-0142-3}}


\bibitem[Xu et~al\mbox{.}(2019)]%
        {xu_adversarial_2019}
\bibfield{author}{\bibinfo{person}{Bing Xu}, \bibinfo{person}{Junfei Zhang},
  \bibinfo{person}{Rui Wang}, \bibinfo{person}{Kun Xu},
  \bibinfo{person}{Yong-Liang Yang}, \bibinfo{person}{Chuan Li}, {and}
  \bibinfo{person}{Rui Tang}.} \bibinfo{year}{2019}\natexlab{}.
\newblock \showarticletitle{Adversarial {Monte} {Carlo} denoising with
  conditioned auxiliary feature modulation}.
\newblock \bibinfo{journal}{\emph{ACM Transactions on Graphics}}
  \bibinfo{volume}{38}, \bibinfo{number}{6} (\bibinfo{date}{Nov.}
  \bibinfo{year}{2019}), \bibinfo{pages}{224:1--224:12}.
\newblock
\showISSN{0730-0301}
\href{https://doi.org/10.1145/3355089.3356547}{doi:\nolinkurl{10.1145/3355089.3356547}}


\bibitem[Xu et~al\mbox{.}(2020)]%
        {xu_unsupervised_2020}
\bibfield{author}{\bibinfo{person}{Zilin Xu}, \bibinfo{person}{Qiang Sun},
  \bibinfo{person}{Lu Wang}, \bibinfo{person}{Yanning Xu}, {and}
  \bibinfo{person}{Beibei Wang}.} \bibinfo{year}{2020}\natexlab{}.
\newblock \showarticletitle{Unsupervised {Image} {Reconstruction} for
  {Gradient}-{Domain} {Volumetric} {Rendering}}.
\newblock \bibinfo{journal}{\emph{Computer Graphics Forum}}
  \bibinfo{volume}{39}, \bibinfo{number}{7} (\bibinfo{year}{2020}),
  \bibinfo{pages}{193--203}.
\newblock
\showISSN{1467-8659}
\href{https://doi.org/10.1111/cgf.14137}{doi:\nolinkurl{10.1111/cgf.14137}}


\bibitem[Yang et~al\mbox{.}(2018)]%
        {yang_fast_2018}
\bibfield{author}{\bibinfo{person}{Xin Yang}, \bibinfo{person}{Dawei Wang},
  \bibinfo{person}{Wenbo Hu}, \bibinfo{person}{Lijing Zhao},
  \bibinfo{person}{Xinglin Piao}, \bibinfo{person}{Dongsheng Zhou},
  \bibinfo{person}{Qiang Zhang}, \bibinfo{person}{Baocai Yin},
  \bibinfo{person}{Qiang Cai}, {and} \bibinfo{person}{Xiaopeng Wei}.}
  \bibinfo{year}{2018}\natexlab{}.
\newblock \showarticletitle{Fast {Reconstruction} for {Monte} {Carlo}
  {Rendering} {Using} {Deep} {Convolutional} {Networks}}.
\newblock \bibinfo{journal}{\emph{IEEE Access}}  \bibinfo{volume}{7}
  (\bibinfo{date}{Dec.} \bibinfo{year}{2018}), \bibinfo{pages}{21177--21187}.
\newblock
\showISSN{2169-3536}
\href{https://doi.org/10.1109/ACCESS.2018.2886005}{doi:\nolinkurl{10.1109/ACCESS.2018.2886005}}


\bibitem[Yang et~al\mbox{.}(2019)]%
        {yang_demc_2019}
\bibfield{author}{\bibinfo{person}{Xin Yang}, \bibinfo{person}{Dawei Wang},
  \bibinfo{person}{Wenbo Hu}, \bibinfo{person}{Li-Jing Zhao},
  \bibinfo{person}{Bao-Cai Yin}, \bibinfo{person}{Qiang Zhang},
  \bibinfo{person}{Xiao-Peng Wei}, {and} \bibinfo{person}{Hongbo Fu}.}
  \bibinfo{year}{2019}\natexlab{}.
\newblock \showarticletitle{{DEMC}: {A} {Deep} {Dual}-{Encoder} {Network} for
  {Denoising} {Monte} {Carlo} {Rendering}}.
\newblock \bibinfo{journal}{\emph{Journal of Computer Science and Technology}}
  \bibinfo{volume}{34} (\bibinfo{date}{Sept.} \bibinfo{year}{2019}),
  \bibinfo{pages}{1123--1135}.
\newblock
\href{https://doi.org/10.1007/s11390-019-1964-2}{doi:\nolinkurl{10.1007/s11390-019-1964-2}}


\bibitem[Yu et~al\mbox{.}(2021)]%
        {yu_monte_2021}
\bibfield{author}{\bibinfo{person}{Jiaqi Yu}, \bibinfo{person}{Yongwei Nie},
  \bibinfo{person}{Chengjiang Long}, \bibinfo{person}{Wenju Xu},
  \bibinfo{person}{Qing Zhang}, {and} \bibinfo{person}{Guiqing Li}.}
  \bibinfo{year}{2021}\natexlab{}.
\newblock \showarticletitle{Monte {Carlo} denoising via auxiliary feature
  guided self-attention}.
\newblock \bibinfo{journal}{\emph{ACM Transactions on Graphics (TOG)}}
  \bibinfo{volume}{40}, \bibinfo{number}{6} (\bibinfo{date}{Dec.}
  \bibinfo{year}{2021}), \bibinfo{pages}{1--13}.
\newblock
\href{https://doi.org/10.1145/3478513.3480565}{doi:\nolinkurl{10.1145/3478513.3480565}}


\bibitem[Zwicker et~al\mbox{.}(2015)]%
        {zwicker_recent_2015}
\bibfield{author}{\bibinfo{person}{Matthias Zwicker}, \bibinfo{person}{Wojciech
  Jarosz}, \bibinfo{person}{Jaakko Lehtinen}, \bibinfo{person}{Bochang Moon},
  \bibinfo{person}{Ravi Ramamoorthi}, \bibinfo{person}{Fabrice Rousselle},
  \bibinfo{person}{Pradeep Sen}, \bibinfo{person}{Cyril Soler}, {and}
  \bibinfo{person}{Sung-Eui Yoon}.} \bibinfo{year}{2015}\natexlab{}.
\newblock \showarticletitle{Recent {Advances} in {Adaptive} {Sampling} and
  {Reconstruction} for {Monte} {Carlo} {Rendering}}.
\newblock \bibinfo{journal}{\emph{Computer Graphics Forum (Proceedings of
  Eurographics - State of the Art Reports)}} \bibinfo{volume}{34},
  \bibinfo{number}{2} (\bibinfo{date}{May} \bibinfo{year}{2015}),
  \bibinfo{pages}{667--681}.
\newblock
\showISSN{1467-8659}
\href{https://doi.org/10.1111/cgf.12592}{doi:\nolinkurl{10.1111/cgf.12592}}


\end{thebibliography}
\balance

\clearpage
\appendix
\section{Appendix}
\subsection{$J(y)$ Functions}
\label{ssec:jensen_gap}

The $J(y)$ functions for all combinations of the loss functions introduced in
Section~\ref{ssec:loss_functions} and the tone mapping functions introduced in
Section~\ref{ssec:model_selection} can be found in
Table~\ref{table:jensen_gap}.  In most cases, since $\varphi'(\hat{y})$ was either convex or concave
as shown by $\varphi'''(\hat{y})$ being either nonnegative or nonpositive, we were able to use
Lemma~1 of \citet{liao_sharpening_2019} to easily calculate the $J(y)$ functions as limits of
$h(x,y)$.  In all of these cases, we had $\lim_{x\to\infty}h(x,y)=0$, and $\lim_{x\to 0}h(x,y)$ gave
the single $J(y)$ function.  However, there were also several cases where $\varphi'(\hat{y})$ was
not convex or concave, and so we had to solve for $\inf_{x\in(0,\infty)}h(x,y)$ and
$\sup_{x\in(0,\infty)}h(x,y)$ directly.  In these cases, there was both a $J_-(y)$ and a $J_+(y)$
function.  In each case, one of these functions was again found at $\lim_{x\to 0}h(x,y)$. For the
other $J(y)$ function, we solved $\frac{\partial}{\partial x}h(x,y)=0$ for $x$ to find the single
critical point, verified that the critical point was the single global minimum or maximum using a
first partial derivative test for pseudoconvexity as in
Section~\ref{ssec:loss_functions}, and then substituted that $x$ value into
$h(x,y)$.  The $\ast$ entries in Table~\ref{table:jensen_gap} represent instances where we were
unable to solve $\frac{\partial}{\partial x}h(x,y)=0$ analytically.  In these instances, we were
still able to find the $J_+(y)$ values numerically, and these results are included in
Figure~\ref{fig:curves}(b).

\subsection{Finite Data}

We now derive a result for finite data similar to the one in Section~3 of the Noise2Noise
Supplemental Material~\citep{lehtinen_noise2noise_2018}, but adapted for our nonlinear Noise2Noise
method.  We have $N$ clean targets $y_i$, $N$ noisy targets $\hat{y}_i$, and $N$ error terms $e_i$
for a finite number $N$.  The $\hat{y}_i$ are random variables defined such that each
$\E[\hat{y}_i]=y_i$, and the $e_i$ are random variables that model the error introduced by the
nonlinearity of the loss function.  The expected squared error between the means of the $y_i$ and
$(\hat{y}_i + e_i)$ terms is~then
\begin{align*}
    &  \E\left[\left(\frac{1}{N}\sum_i y_i - \frac{1}{N}\sum_i(\hat{y}_i + e_i)\right)^2\right] \\
    &= \frac{1}{N^2}\left(\left(\sum_i y_i\right)^2 - 2\left(\sum_i y_i\right)\E\left[\sum_i(\hat{y}_i + e_i)\right] + \E\left[\left(\sum_i(\hat{y}_i + e_i)\right)^2\right] \right) \\
    &= \frac{1}{N^2}\left(\left(\sum_i y_i\right)^2 - 2\left(\sum_i y_i\right)\E\left[\sum_i \hat{y}_i\right] - 2\left(\sum_i y_i\right)\E\left[\sum_i e_i\right] \right. \\ &\qquad \left.
        + \E\left[\left(\sum_i\hat{y}_i\right)^2\right] + 2\E\left[\sum_i\hat{y}_i\sum_i e_i\right]  + \E\left[\left(\sum_i e_i\right)^2\right] \right) \\
    &= \frac{1}{N^2}\left(\Var\left(\sum_i\hat{y}_i\right) + 2\Cov\left(\sum_i\hat{y}_i\;,\sum_i e_i\right) \right. \\ &\qquad \left. + \Var\left(\sum_i e_i\right) + \E\left[\sum_i e_i\right]^2 \right) \\
    &= \frac{1}{N^2}\left(\Var\left(\sum_i(\hat{y}_i + e_i)\right) + \E\left[\sum_i e_i\right]^2 \right) \displaybreak[0] \\
    &= \frac{1}{N}\left(\frac{1}{N}\sum_i\sum_j\Cov(\hat{y}_i + e_i\,,\hat{y}_j + e_j)\right) + \left(\frac{1}{N}\sum_i\E[e_i]\right)^2 \\
    &= \frac{1}{N}\left(\frac{1}{N}\sum_i\Var(\hat{y}_i + e_i)\right) + \left(\frac{1}{N}\sum_i\E[e_i]\right)^2 \,,
\end{align*}
where the last line is a simplification that is valid when the $(\hat{y}_i + e_i)$ terms are
uncorrelated.  We can see from the last two lines above that the expected error is composed of two
overall terms: a variance term and a bias term.  The error from training with noisy targets
$\hat{y}_i$ occurs entirely in the variance term, because the noisy targets have the correct mean
$\E[\hat{y}_i]=y_i$.  This average variance term decreases as the number of samples increases.  The
error from the nonlinearity of the loss function $e_i$ occurs in both the variance and bias terms.
The variance component decreases as the number of samples increases, while the bias component does
not vary with the number of samples.

\subsection{Additional Qualitative Results}
\label{ssec:qualitative}

\begin{figure*}[p]
    \centering
    \includegraphics[width=\textwidth]{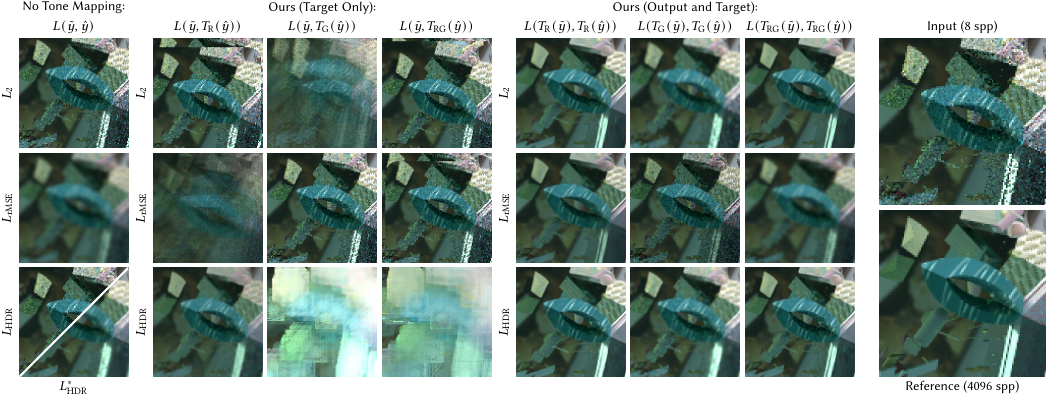} \\ \vspace{0.5cm}
    \includegraphics[width=\textwidth]{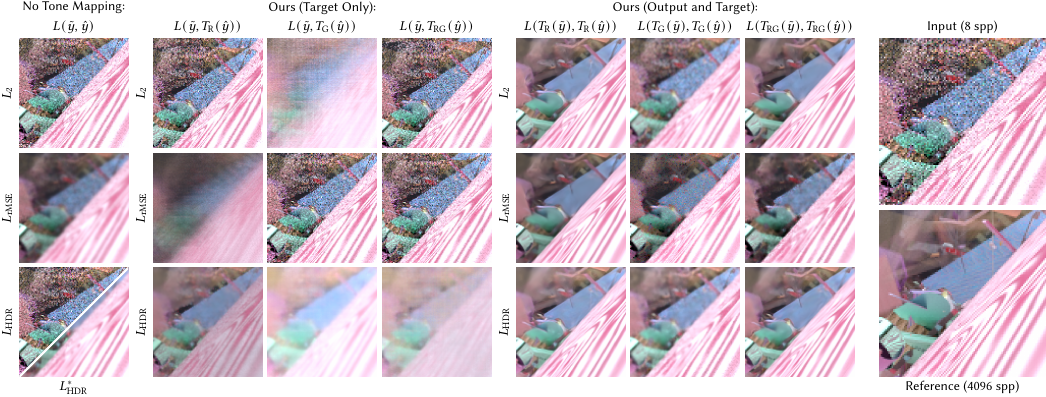}
    \caption{Model outputs for a single validation image for all combinations of loss functions,
    tone mapping functions, and tone mapping placement}
    \label{fig:grid_extra}
    \Description{Two grids of images, each with 3 rows and 7 columns.  The rows are labeled
    L-2, L-rMSE, and L-HDR.  The loss functions in the first column have no tone mapping, and the
    bottom image in this column is split in half diagonally with the lower half showing the output
    for L-HDR Star.  The next three columns have tone mapping on the target image only for T-R, T-G,
    and T-RG, respectively.  The last three columns have tone mapping on the model output and target
    images, again for the T-R, T-G, and T-RG tone mapping functions.  The quality of the outputs
    generally improves from left to right, with some outputs towards the right still displaying some
    blurring and noise.  To the right of the main grids are two additional images showing the noisy
    input and clean reference images.}
\end{figure*}

\begin{figure*}[p]
    \centering
    \includegraphics[width=\textwidth]{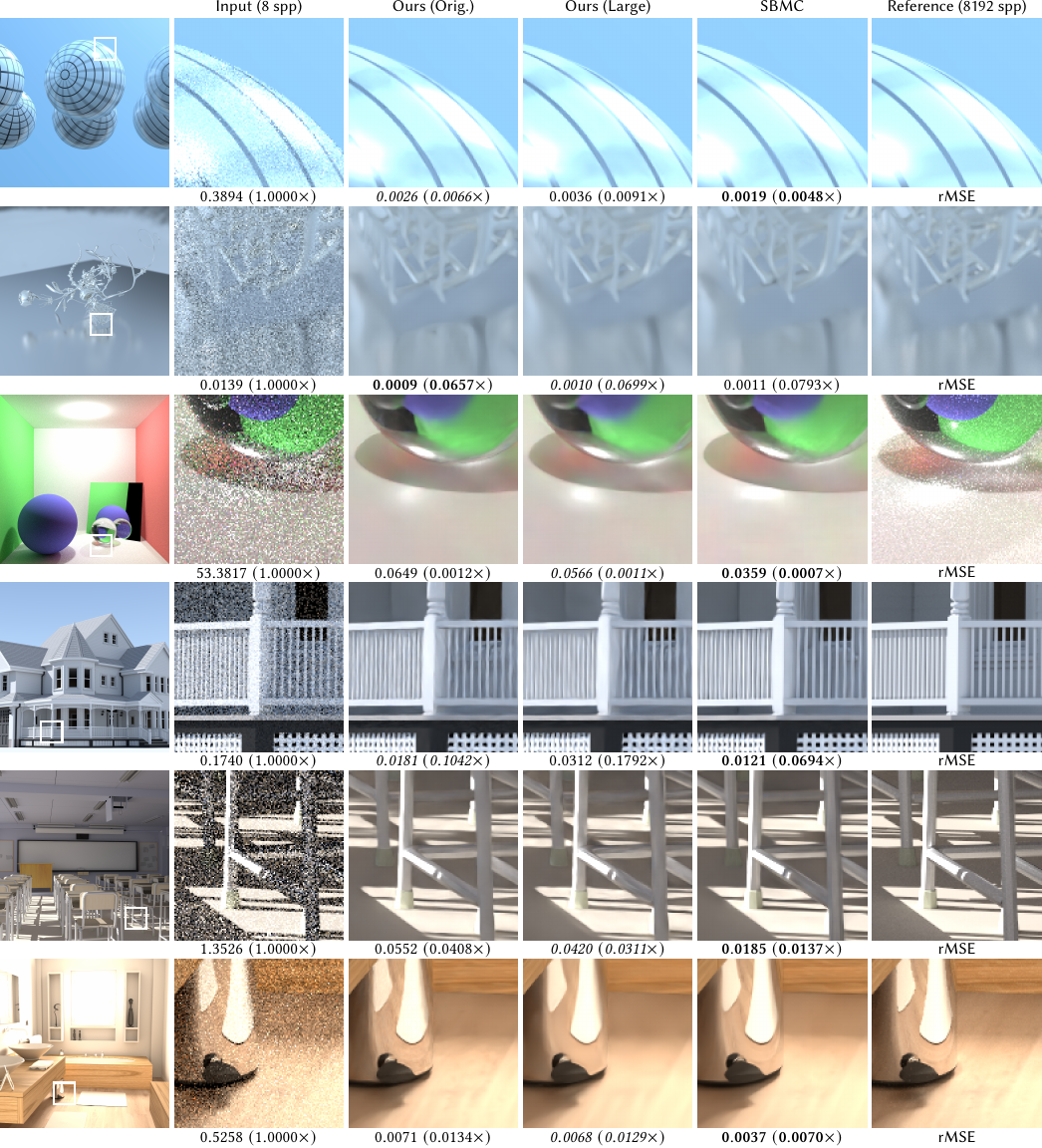}
    \caption{Comparison between the outputs of our denoising models and \acrshort{sbmc} on several
    test images. Columns from left to right: overview of the test image with the crop region
    outlined, noisy input rendered at 8~\acrfull{spp}, output of our original denoising model,
    output of our large denoising model, \acrshort{sbmc} output, and reference image rendered at
    8192~\acrshort{spp}. (\href{https://pbrt.org/scenes-v2}{anim-bluespheres} Copyright \copyright
    1998-2012, \href{https://pbrt.org/}{Matt Pharr and Greg Humphreys};
    \href{https://www.cs.cmu.edu/~kmcrane/Projects/ModelRepository/}{Yeah Right} by
    \href{https://www.cs.cmu.edu/~kmcrane/index.html}{Keenan Crane},
    \href{https://creativecommons.org/publicdomain/zero/1.0/}{CC0~1.0};
    \href{https://github.com/adobe/sbmc}{GITestSynthesizer\_01} Copyright \copyright 2019,
    \href{https://mgharbi.com/}{Michaël Gharbi};
    \href{https://blendswap.com/blend/12687}{Victorian Style House} by
    \href{https://blendswap.com/profile/10069}{MrChimp2313},
    \href{https://creativecommons.org/publicdomain/zero/1.0/}{CC0~1.0};
    \href{https://blendswap.com/blend/13632}{Japanese Classroom} by
    \href{https://blendswap.com/profile/135376}{NovaAshbell},
    \href{https://creativecommons.org/licenses/by/3.0/}{CC~BY~3.0};
    \href{https://blendswap.com/blend/12584}{Bathroom} by
    \href{https://blendswap.com/profile/72536}{nacimus},
    \href{https://creativecommons.org/licenses/by/3.0/}{CC~BY~3.0})}
    \label{fig:comparison}
    \Description{Six rows of images with six images in each row.  The first image in each
    row is an overview of a rendered scene with a crop region highlighted. The second image in each
    row is a noisy rendering of the cropped region. The third and fourth images in each row are the
    outputs of our original and large models, respectively. The fifth image in each row is the
    output of the SBMC model.  The final image in each row is a high-sample count reference image.
    The outputs of our models and SBMC are generally similar, with slight differences in certain
    areas (see text).}
\end{figure*}

Figure~\ref{fig:grid_extra} shows some additional copies of
Figure~\ref{fig:tone_loss_images} featuring different validation images.  These
images provide further evidence for the findings in
Section~\ref{ssec:model_selection}.  Figure~\ref{fig:comparison} shows some
additional outputs from our models and \gls{sbmc} on example images from the \gls{sbmc} test set.
The first scene involves rotating spheres with motion blur hovering over a mirror.  The zoomed-in
portion shows one of the spheres with specular highlights and blurred textures, where the outputs
from our models are similar to that of \gls{sbmc} and to the reference image.  The second scene
contains a complicated metallic structure that is suspended over a reflective surface.  The
zoomed-in portion shows that our models again produce similar output to \gls{sbmc}, capturing the
details of the structure but missing some small areas of shadow.  Both our models and \gls{sbmc}
over-blur some of the reflection of the structure.  The third scene shows a Cornell box with matte,
reflective, and refractive objects.  The zoomed-in portion shows that the reference image is still
quite noisy.  All models produce outputs that are visually better than the reference image, with our
models producing smaller caustics than \gls{sbmc}.  Both our models and \gls{sbmc} over-blur the
reflection of the purple sphere and miss a highlight on the glass sphere.  The fourth scene features
a house model in outdoor lighting conditions.  The zoomed-in portion shows that our models over-blur
the railing, producing outputs that are less sharp than \gls{sbmc}.  Our models also over-blur the
window frame above the railing, although all of the models distort the part of the window frame that
is visible through the railing.  The fifth scene shows a classroom with light streaming in through
the windows.  The zoomed-in portion shows that the outputs are similar between our models and
\gls{sbmc}, although our models slightly over-blur the chair feet and the point of the shadow on the
floor.  The final scene is a bathroom lit by a diffuse light source on the far wall.  The zoomed-in
portion shows that our models produce similar output to \gls{sbmc}, but miss some detail in the wood
texture and in the reflection of the floor in the trash can.  Both our models and \gls{sbmc}
over-blur the floor texture where it is in shadow.

\makeatletter
\setlength{\@dblfptop}{0pt plus 1fil}
\setlength{\@dblfpbot}{0pt plus 1fil}
\makeatother
\begin{table*}[p]
    \centering
    \caption{$J(y)$ functions for various combinations of tone mapping and loss functions.  A blank
    space means that the function is $J(y)=0$.  See Section~\ref{ssec:jensen_gap} for details on the
    $\ast$ values.}
    \makegapedcells
    \begin{tabular}{ccc}\hline
        $\varphi(\hat{y})$ & $J_-(y)$ & $J_+(y)$ \\ \hline
        $T_\mathrm{R}=\frac{\hat{y}}{1+\hat{y}}$ & $- \frac{1}{\left(y + 1\right)^{2}}$ & \\ \hline
        $T_\mathrm{G}=\hat{y}^\frac{1}{2.2}$ & $- \frac{6}{11 y^{\frac{17}{11}}}$ & \\ \hline
        $T_\mathrm{RG}=T_\mathrm{G}\circ T_\mathrm{R}$ & $- \frac{11 y + 6}{11 y^{\frac{17}{11}} \left(y + 1\right)^{\frac{16}{11}}}$ & \\ \hline
        $\frac{1}{(\hat{y}+\epsilon)^2}$ & & $\frac{3 \epsilon + y}{\epsilon^{2} \left(\epsilon + y\right)^{3}}$ \\ \hline
        $\frac{\hat{y}}{(\hat{y}+\epsilon)^2}$ & $- \frac{2}{\left(\epsilon + y\right)^{3}}$ & $\frac{\left(- \epsilon + y\right)^{2}}{4 \epsilon \left(\epsilon + y\right)^{4}}$ \\ \hline
        $\frac{1}{(T_\mathrm{R}+\epsilon)^2}$ & & $\frac{2 \epsilon^{2} y + 2 \epsilon^{2} + 3 \epsilon y + 3 \epsilon + y}{\epsilon^{2} \left(\epsilon y + \epsilon + y\right)^{3}}$ \\ \hline
        $\frac{T_\mathrm{R}}{(T_\mathrm{R}+\epsilon)^2}$ & $- \frac{\epsilon y + \epsilon + y + 2}{\left(\epsilon y + \epsilon + y\right)^{3}}$ & $\frac{\left(\epsilon + 1\right)^{2} \left(\epsilon y + \epsilon - y\right)^{2}}{4 \epsilon \left(\epsilon y + \epsilon + y\right)^{4}}$ \\ \hline
        $\frac{1}{(T_\mathrm{G}+\epsilon)^2}$ & & $\frac{12 \epsilon^{2} y + 33 \epsilon y^{\frac{16}{11}} + 11 y^{\frac{21}{11}}}{11 \epsilon^{2} y^{\frac{28}{11}} \left(\epsilon + y^{\frac{5}{11}}\right)^{3}}$ \\ \hline
        $\frac{T_\mathrm{G}}{(T_\mathrm{G}+\epsilon)^2}$ & $- \frac{2 \left(3 \epsilon y + 8 y^{\frac{16}{11}}\right)}{11 y^{\frac{28}{11}} \left(\epsilon + y^{\frac{5}{11}}\right)^{3}}$ & $\ast$ \\ \hline
        $\frac{1}{(T_\mathrm{RG}+\epsilon)^2}$ & & $\frac{22 \epsilon^{2} y^{2} + 12 \epsilon^{2} y + 33 \epsilon y^{\frac{16}{11}} \left(y + 1\right)^{\frac{6}{11}} + 11 y^{\frac{21}{11}} \left(y + 1\right)^{\frac{1}{11}}}{11 \epsilon^{2} y^{\frac{28}{11}} \left(y + 1\right)^{\frac{1}{11}} \left(\epsilon \left(y + 1\right)^{\frac{5}{11}} + y^{\frac{5}{11}}\right)^{3}}$ \\ \hline
        $\frac{T_\mathrm{RG}}{(T_\mathrm{RG}+\epsilon)^2}$ & $- \frac{11 \epsilon y^{2} \left(y + 1\right)^{\frac{5}{11}} + 6 \epsilon y \left(y + 1\right)^{\frac{5}{11}} + 11 y^{\frac{27}{11}} + 16 y^{\frac{16}{11}}}{11 y^{\frac{28}{11}} \left(y + 1\right)^{\frac{6}{11}} \left(\epsilon \left(y + 1\right)^{\frac{5}{11}} + y^{\frac{5}{11}}\right)^{3}}$ & $\ast$ \\ \hline
        $T_\mathrm{R}^2$ & $- \frac{y^{2}}{\left(y + 1\right)^{4}}$ & $\begin{cases}\frac{1 - y}{(y + 1)^3} & 0\leq y\leq1 \\\quad 0 & 1\leq y\end{cases}$ \\ \hline
        $T_\mathrm{G}^2$ & $- \frac{1}{11 y^{\frac{12}{11}}}$ & \\ \hline
        $T_\mathrm{RG}^2$ & $- \frac{11 y + 1}{11 y^{\frac{12}{11}} \left(y + 1\right)^{\frac{21}{11}}}$ & \\ \hline
    \end{tabular}
    \label{table:jensen_gap}
\end{table*}

\end{document}